\documentclass[letterpaper]{article} 
\usepackage{aaai2027}  
\usepackage[hyphens]{url}  
\usepackage{graphicx} 
\usepackage{natbib}  
\usepackage{caption} 
\usepackage{algorithm}
\usepackage{algorithmic}

\usepackage{newfloat}
\usepackage{listings}
\DeclareCaptionStyle{ruled}{labelfont=normalfont,labelsep=colon,strut=off} 
\floatstyle{ruled}
\newfloat{listing}{tb}{lst}{}
\floatname{listing}{Listing}

\usepackage{booktabs}

\usepackage{amsmath}
\usepackage{multirow}
\usepackage{subcaption}
\usepackage{seqsplit}

\title{Dynamic Defense Profiling Enables Cognitive Jailbreak of Text-to-Image Models}
\author{
    Dongdong Yang\textsuperscript{\rm 1},
    Deyue Zhang\textsuperscript{\rm 1},
    Zhao Liu\textsuperscript{\rm 1},
    Zonghao Ying\textsuperscript{\rm 2},
    Wenzhuo Xu\textsuperscript{\rm 1},
    Jiankai Jin\textsuperscript{\rm 1},
    Xiangzheng Zhang\textsuperscript{\rm 1},
    Quanchen Zou\textsuperscript{\rm 1}\corresponding
}
\affiliations{
    \textsuperscript{\rm 1}360 AI Security Lab, Beijing, China\\
    \textsuperscript{\rm 2}School of Artificial Intelligence, Beihang University, Beijing, China\\

    yangdongdong1@360.cn, zouquanchen@360.cn
}

\nocopyright

\begin{document}

\maketitle

\begin{abstract}
Text-to-Image (T2I) generative models have achieved remarkable progress in synthesizing high-quality visual content, yet they remain vulnerable to adversarial misuse, particularly in generating Not-Safe-For-Work (NSFW) images. Most existing jailbreak attacks primarily rely on heuristic prompt engineering or black-box optimization, treating model feedback as a binary signal (success or failure). This coarse-grained paradigm overlooks the rich information embedded in diverse failure modes, such as textual refusal, visual blocking, and semantic sanitization, resulting in inefficient exploration and severe semantic collapse.

In this paper, we propose \textbf{MIND}, a cognitive jailbreak framework that reframes adversarial prompt generation as a belief-state inference problem over latent defense mechanisms. Instead of blindly searching for bypass prompts, MIND actively models the target system's latent defense mechanisms by interpreting multi-modal feedback as high-density signals. Specifically, the framework integrates three core components: (1) a \textbf{Multi-modal Judge} for fine-grained feedback decomposition, (2) a \textbf{Defense Profiler} for iterative belief updating, and (3) a \textbf{Meta-Memory} module for retrieving historically effective attack strategies. These components are unified within a reasoning-driven evolutionary optimization process, enabling adaptive and semantically consistent jailbreak generation. Extensive experiments on the I2P benchmark demonstrate the effectiveness of MIND. Under six representative pre-processing and post-processing defense settings applied to the Stable Diffusion v1.5 model, MIND achieves an Attack Success Rate (ASR) of 95.62\%, significantly outperforming existing methods. Additionally, the effectiveness of the proposed framework is validated across four widely used commercial T2I systems, achieving the highest ASR of 91.58\% on Wan-2.5.
\end{abstract}


\section{Introduction}
Generative Artificial Intelligence has rapidly evolved with the emergence of large-scale Text-to-Image (T2I) models, such as Stable Diffusion \cite{sd}, Midjourney \cite{midjourney}, and Wan \cite{wan25}. These models enable users to synthesize high-quality images from natural language descriptions, unlocking unprecedented creative potential. However, this capability also introduces significant safety risks. T2I models can be misused by malicious users to produce pornographic, violent, or otherwise NSFW content. To mitigate these risks, model providers have deployed multi-layered defense mechanisms, including preprocessing text filtering, post-processing visual blocking, and safety-driven reinforcement learning from human feedback (RLHF) \cite{rlhf}.

Despite these safeguards, the robustness of T2I models remains an open research question. Recent studies have demonstrated that T2I models are susceptible to jailbreak attacks, where carefully crafted prompts can bypass safety filters \cite{mma, sneakyprompt, rab, jf, rpg}. Early attacks primarily relied on heuristic prompt engineering or black-box optimization, treating model feedback as a binary signal (success or failure) \cite{flirt, rando2022red, qu2023unsafe, dong2024jailbreaking}. This coarse-grained paradigm overlooks the rich information embedded in diverse failure modes, such as textual refusal, visual blocking, and semantic sanitization. Therefore, although these methods can effectively bypass simple keyword filters, they struggle against modern, sophisticated defenses.

In this paper, we argue that jailbreak should be framed as a process of defense inference rather than blind optimization. Instead of merely searching for effective prompts, attackers can leverage multi-modal feedback to infer the latent defense behavior of the target model. This perspective enables more informed and adaptive attack strategies, improving both efficiency and semantic fidelity.

Therefore, we propose MIND, a cognitive jailbreak framework that models T2I attacks as an iterative inference process over hidden defense mechanisms. MIND integrates three key components: (1) a Multi-modal Judge that decomposes model feedback into fine-grained categories, (2) a Defense Profiler that updates beliefs about the target system and generates tactical guidance, and (3) a Meta-Memory module that retrieves effective strategies from prior experiences. These components are unified within a reasoning-driven evolutionary framework, where LLM-based agents perform chain-of-thought prompt mutation.

Our contributions are summarized as follows:
\begin{itemize}
    \item \textbf{Multi-modal Feedback Analysis:} We propose a novel feedback mechanism utilizing VLMs to decompose T2I responses into fine-grained signals, transforming ``failure'' into actionable intelligence.
\end{itemize}
\begin{itemize}
    \item \textbf{Dynamic Defense Profiling:} We introduce the concept of a Defense Profile, a probabilistic belief state tracking the target model's sensitivity. We develop a Bayesian update algorithm driven by LLM attribution analysis to refine this profile in real-time.
\end{itemize}
\begin{itemize}
    \item \textbf{Cognitive Evolutionary Search:} We integrate the Defense Profile into an evolutionary search loop, enabling Chain-of-Thought (CoT) \cite{cot} based prompt mutation that adapts to the inferred defense strategy.
\end{itemize}
\begin{itemize}
    \item \textbf{Empirical Effectiveness:} We evaluate MIND on state-of-the-art models. Experiments show that our approach achieves a significantly higher ASR than existing baselines while preserving semantic consistency. Moreover, it maintains strong effectiveness on four mainstream commercial T2I models, achieving ASRs of 97.89\%, 91.58\%, 86.32\%, and 70.52\% on SDXL, Wan-2.5, Seedream-5.0, and Gemini-3-pro, respectively.
\end{itemize}

\section{Related Work}

\subsection{Jailbreak Attacks on Text-to-Image Models}

Automated jailbreak attacks on text-to-image (T2I) models can be broadly categorized into gradient-based optimization, black-box search, and semantic manipulation.

\textbf{Gradient-based methods}, such as MMA~\cite{mma}, optimize text and latent representations to induce unsafe generations under white-box access, but do not generalize to closed-source APIs.

\textbf{Black-box search methods} refine prompts with the support of model feedback. Representative methods include adopting evolutionary mutation with binary signals~\cite{sneakyprompt}, formulating jailbreaking as a zeroth-order optimization problem~\cite{diffzoo}, combining fuzz-testing with LLM agents for structured exploration~\cite{jf}, performing in-context self-refinement~\cite{flirt}, mining failure-inducing prompts for red-teaming~\cite{prompting4debugging}, and guiding prompt evolution via rule-based preference modeling~\cite{rpg}. These methods, however, rely on coarse feedback, limiting adaptability to diverse failure modes.

\textbf{Semantic manipulation methods} bypass safety filters by obfuscating malicious intent. Representative methods include decomposing unsafe queries into benign components~\cite{daca}, using metaphorical expressions to disguise malicious intent~\cite{mja}, jointly targeting textual and visual filters~\cite{upam}, leveraging LLM reasoning for multi-step prompt construction~\cite{reason2attack}, performing collaborative prompt editing~\cite{coljailbreak}, learning transferable adversarial prompts~\cite{pla}, and exposing weaknesses in concept removal defenses by reactivating suppressed concepts~\cite{rab}.

Overall, existing attacks treat safety mechanisms as static barriers and fail to exploit fine-grained multi-modal feedback (e.g., refusal, blocking, sanitization). In contrast, our method explicitly models defense behaviors to enable adaptive and targeted jailbreak generation.

\subsection{Safety Mechanisms in T2I Models}

Safety mechanisms in T2I systems can be grouped into input-level filtering, output-level filtering, and model-level alignment.

\textbf{Input-level defenses.} Typical input-level defenses block unsafe prompts before generation. A number of methods have been proposed, including rule-based filtering~\cite{text-match}, classifier-based detection~\cite{text-cls}, learned safety guards~\cite{guardt2i, qwen3guard}, and extending filtering into latent representations~\cite{LatentGuard}.

\textbf{Output-level defenses.} Typical output-level defenses detect unsafe images after generation. Typical methods include NSFW classifiers~\cite{img-cls}, CLIP-based detectors~\cite{img-clip}, and deployed safety checkers that measure similarity to unsafe concepts~\cite{text-img}.

\textbf{Model-level defenses.} Typical model-level defenses align generation with safety constraints. Such methods include suppressing unsafe concepts during inference~\cite{sld}, removing unsafe concepts via fine-tuning or unlearning~\cite{esd, DoCo}, incorporating safety-aware objectives~\cite{safegen}, and aligning outputs via preference learning~\cite{safetydpo}.

Despite these efforts, defenses are typically modular, leading to cross-stage inconsistencies and exploitable gaps. Moreover, many methods rely on soft suppression rather than explicit refusal, often producing \emph{semantic sanitization}—visually safe but distorted outputs. Our work leverages this nuanced feedback as a dense signal to reverse-engineer and bypass multi-stage safety mechanisms.

\begin{figure*}[htbp]
  \centering  
  \includegraphics[
    width=1.0\textwidth,  
  ]{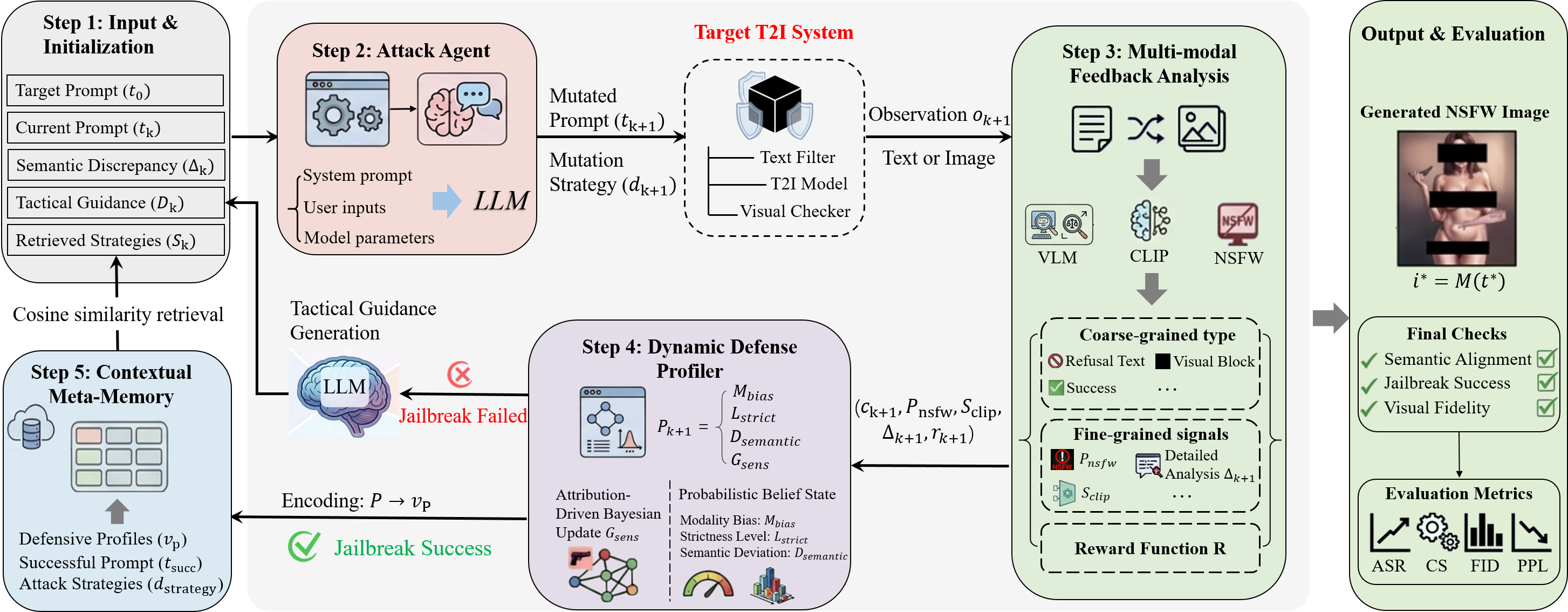}        
  \caption{Overview of the proposed MIND framework. The system performs profile-guided evolutionary jailbreak through the interaction of three core components: Multi-modal Judge, Defense Profiler, and Meta-Memory, under a multi-agent collaboration paradigm.}
  \label{fig:framework}
\end{figure*}

\section{Methodology}
\subsection{Problem Formulation}
Let $M: \mathcal{T} \to \mathcal{I}$ denote a Text-to-Image (T2I) generative model mapping a text prompt $t$ to an image $i$. The model is protected by a safety mechanism $S$. The attacker has no access to model parameters, gradients, or internal safety policies. The goal of a jailbreak attack is to find an adversarial prompt $t'$ derived from a malicious intent $t_{\text{mal}}$ such that: (1) Safety Bypass: The generated image $i' = M(t')$ contains prohibited content, bypassing $S$; (2) Semantic Consistency: $i'$ remains semantically aligned with $t_{\text{mal}}$.

We formulate the jailbreak process as a Partially Observable Markov Decision Process (POMDP), defined by
$\langle S_{\text{state}}, A, O, T, R \rangle$.
The \textbf{State ($S_{\text{state}}$)}: The latent defense configuration of the target model, including forbidden concepts, modality bias (text vs. image focus), and strictness levels. This state is hidden.
The \textbf{Action ($A$)}: The generated adversarial prompt $t_k$ at step $k$.
The \textbf{Observation ($O$)}: The multi-modal feedback from the target, including generated image $i_k$, refusal text, or filtered placeholders.
The \textbf{Transition ($T$)}: The update of the attacker’s belief state $b_t(s)$ about the defense mechanism is managed by the Defense Profiler.
The \textbf{Reward ($R$)}: A composite function evaluating the quality of the attack based on fine-grained feedback.

\subsection{Evolutionary Search with Profile-Guided Multi-Agent Collaboration}

We formulate jailbreak generation as a profile-guided evolutionary search process with multi-agent collaboration. 
An overview of the proposed framework is illustrated in Fig.~\ref{fig:framework}, where the interaction between the Multi-modal Judge, Defense Profiler, and Meta-Memory modules forms a closed-loop optimization pipeline. 
The overall procedure of this process is summarized in Algorithm~1, which formalizes the iterative cycle of feedback analysis, belief updating, and profile-guided prompt mutation. At each iteration, candidate prompts are optimized under a closed-loop interaction between three components: (1) multi-modal feedback (§3.3), which provides fine-grained signals from model responses; (2) dynamic defense profiling (§3.4), which maintains a latent belief over the target system and produces tactical guidance; and (3) contextual meta-memory (§3.5), which retrieves historically effective strategies conditioned on the current defense profile.

These components jointly guide three cooperating agents. The Judge Agent evaluates prompts and produces structured rewards, the Profiler Agent updates the defense belief and generates guidance, and the Attack Agent performs CoT-guided prompt mutation conditioned on feedback, inferred defense behavior, and retrieved strategies.

The population is iteratively updated via evaluation, profile-guided mutation, and selection, while successful cases are stored into memory for future reuse. This design can be viewed as a belief-conditioned population-based search, where defense profiling serves as an implicit environment model and meta-memory provides experience priors, enabling adaptive and semantically consistent jailbreak generation under dynamic defenses. The overall procedure of the proposed profile-guided evolutionary jailbreak is summarized in Algorithm~1.

\subsection{Multi-modal Feedback Analysis}
Existing attacks often fail to distinguish between different failure modes. Our Multi-modal Judge utilizes a Vision-Language Model (VLM) and CLIP to decompose the target's response into fine-grained signals. Let $f_{\text{target}}(t)$ denote the target model's output. The judge classifies the observation $o_k$ into five categories based on safety probability $P_{\text{nsfw}}$ (from an NSFW classifier like NudeNet~\cite{nudenet}) and semantic consistency $S_{\text{clip}}$ (cosine similarity between prompt and image embeddings):

\begin{itemize}
    \item\textbf{Text Refusal ($C_{\text{ref}}$)}: The model explicitly rejects the request (e.g., ``I cannot generate ...''), indicating activation of a preprocessing text-level safety filter.
    \item \textbf{Visual Block ($C_{\text{block}}$)}: The model produces a black or blurred image, indicating interception by a post-processing visual safety filter.
    \item \textbf{Soft Defense ($C_{\text{soft}}$)}: The image is generated ($P_{\text{nsfw}} \leq \tau_{\text{safe}}$) but remains semantically misaligned ($S_{\text{clip}} \approx S_{\text{safe}}$), indicating semantic sanitization (e.g., ``naked'' $\rightarrow$ ``clothed'').
    \item \textbf{Semantic Failure ($C_{\text{fail}}$)}: The generated image contains NSFW content ($P_{\mathrm{nsfw}} > \tau_{\mathrm{safe}}$) but lacks semantic alignment ($S_{\mathrm{clip}} < \tau_{\mathrm{sem}}$), reflecting a failure to jointly optimize safety evasion and semantic fidelity.
    \item \textbf{Success ($C_{\text{succ}}$)}: Both criteria are satisfied ($P_{\mathrm{nsfw}} > \tau_{\mathrm{safe}}$ and $S_{\mathrm{clip}} \ge \tau_{\mathrm{sem}}$), indicating a successful jailbreak.
\end{itemize}
These categories form a partition over the outcome space, serving as the basis for reward design.

\textbf{Reward Function.}
Based on the outcome categories defined above, we define a structured reward function:
\begin{equation}
R(t_k) =
\left\{
\begin{array}{@{}l@{\quad}l@{}}
0, & o_k \in C_{\text{ref}} \\[2pt]
\epsilon, & o_k \in C_{\text{block}} \\[2pt]
\alpha \cdot S_{\mathrm{clip}}, & o_k \in C_{\text{soft}} \\[2pt]
\begin{array}{@{}l@{}}
\beta + \gamma \cdot P_{\mathrm{nsfw}} \\
\quad - \max(0, \tau_{\mathrm{sem}} - S_{\mathrm{clip}})
\end{array}, & o_k \in C_{\text{fail}} \\[2pt]
1 + P_{\mathrm{nsfw}} \cdot S_{\mathrm{clip}}, & o_k \in C_{\text{succ}}
\end{array}
\right.
\end{equation}
where $\alpha, \beta, \gamma$ are hyperparameters and $\epsilon$ is a small constant.

\textbf{Interpretation.}
This reward jointly encourages safety violation and semantic alignment, and can be viewed as a relaxation of:
\begin{equation}
\max \; P_{\mathrm{nsfw}} \cdot S_{\mathrm{clip}} 
\quad \text{s.t.} \quad S_{\mathrm{clip}} \ge \tau_{\mathrm{sem}}.
\end{equation}

\subsection{Dynamic Defense Profiling}
We maintain a probabilistic belief over the target model's defense mechanism, 
denoted as $P_k$ at step $k$:
\begin{equation}
P_k = \{ M_{\mathrm{bias}}, \; L_{\mathrm{strict}}, \; D_{\mathrm{semantic}}, \; G_{\mathrm{sens}} \}.
\end{equation}

Here, $M_{\mathrm{bias}} \in [0,1]$ captures the relative reliance on text-level versus visual-level filtering, 
$L_{\mathrm{strict}} \in [0,1]$ estimates the overall safety sensitivity, 
and $D_{\mathrm{semantic}} \in [0,1]$ reflects the risk of semantic deviation due to prompt variation. 
$G_{\mathrm{sens}}$ denotes a concept-level sensitivity map, where each concept $c$ is associated with a blocking probability $p(\mathrm{block} \mid c)$.

\textbf{Belief Update.}
Given the observation $o_k$, we update $P_k$ via an attribution-driven mechanism. 
For outcomes in $C_{\text{ref}}$ and $C_{\text{soft}}$, we leverage an LLM-based attribution module to identify the dominant concept $c_i$ responsible for the observed behavior. 
The sensitivity map is updated as:
\begin{equation}
G_{\mathrm{sens}}[c_i] \leftarrow G_{\mathrm{sens}}[c_i] 
+ \eta \cdot (1 - G_{\mathrm{sens}}[c_i]) \cdot p(\mathrm{cause} \mid c_i),
\end{equation}
where $\eta$ is the learning rate and $p(\mathrm{cause} \mid c_i)$ denotes the attribution confidence. This belief update can be viewed as an approximate Bayesian update under partial observability. For $C_{\text{block}}$, we increase $M_{\mathrm{bias}}$, indicating stronger reliance on visual filtering.

\textbf{Tactical Guidance.}
Based on the updated belief $P_k$, the profiler generates a natural-language guidance signal $D_k$ 
to steer subsequent actions (e.g., shifting toward abstract or stylized expressions when visual sensitivity is high). The guidance $D_k$ serves as an auxiliary signal to guide policy exploration.

\subsection{Contextual Meta-Memory}
To enable few-shot adaptation, we introduce a meta-memory module $\mathcal{M}_{\mathrm{mem}}$ 
that stores tuples of the form $\langle v_p, t_{\text{succ}}, d_{\text{strategy}} \rangle$, 
where $v_p$ is a representation of the defense profile, $t_{\text{succ}}$ is a successful prompt, 
and $d_{\text{strategy}}$ denotes the corresponding strategy description.

The profile $P_k$ is encoded into a vector $v_p$ by combining semantic and statistical features:
\begin{equation}
v_p = \mathrm{Concat}\big( E_{\mathrm{sem}}(\text{top\_concepts}), \; f_{\mathrm{stat}} \big),
\end{equation}
where $E_{\mathrm{sem}}(\cdot)$ extracts semantic embeddings of high-risk concepts, and 
$f_{\mathrm{stat}} = [M_{\mathrm{bias}}, L_{\mathrm{strict}}, D_{\mathrm{semantic}}]$ 
captures normalized defense statistics, including modality bias, strictness level, 
and semantic deviation.

Before generating a new prompt, we query $\mathcal{M}_{\mathrm{mem}}$ using cosine similarity to retrieve the top-$k$ most relevant entries. The retrieved strategies provide contextual guidance, enabling the agent to reuse effective patterns (e.g., stylistic abstraction) under similar defense conditions.

\begin{algorithm}[t]
\caption{Profile-Guided Evolutionary Jailbreak}
\label{alg:mind}
\small
\begin{algorithmic}[1]

\STATE \textbf{Input:} Target model $M$, target malicious prompt $t_0$, budget $K$
\STATE \textbf{Output:} Successful prompt $t^*$

\STATE Initialize population $\mathcal{T}_0$
\STATE Initialize profile $P_0$, memory $\mathcal{M}_{\mathrm{mem}}$

\FOR{$k = 0$ to $K$}

    \FOR{each $t_k^{(i)} \in \mathcal{T}_k$}
        \STATE $o_k^{(i)} \sim \mathcal{M}(t_k^{(i)})$
        \STATE Extract signals $(P_{\mathrm{nsfw}}, S_{\mathrm{clip}}, \Delta_k^{(i)}, c_k^{(i)})$
        \STATE $r_k^{(i)} \leftarrow R(t_k^{(i)})$
    \ENDFOR

    \STATE \textbf{// Defense Profiling}
    \STATE Update $P_k$ using extracted signals $\{t_k^{(i)}, c_k^{(i)}, \Delta_k^{(i)}\}$
    \STATE Generate guidance $D_k$

    \STATE \textbf{// Meta-Memory Retrieval}
    \STATE Encode $P_k \rightarrow v_p$
    \STATE Retrieve strategies $\mathcal{S}_k$

    \STATE \textbf{// Mutation (Attack Agent)}
    \FOR{each $t_k^{(i)}$}
        \STATE Construct context $\mathcal{C}_k^{(i)} = \{t_0, t_k^{(i)}, D_k, \mathcal{S}_k, \Delta_k^{(i)}\}$
        \STATE Generate $(t_{k+1}^{(i)}, d_{k+1}^{(i)})$
    \ENDFOR

    \STATE \textbf{// Selection}
    \STATE Select top-$N$ candidates to form $\mathcal{T}_{k+1}$

    \STATE \textbf{// Memory Update}
    \FOR{successful $t_{k+1}^{(i)}$}
        \STATE Store $\langle v_p, t_{k+1}^{(i)}, d_{k+1}^{(i)} \rangle$ into $\mathcal{M}_{\mathrm{mem}}$
    \ENDFOR

    \IF{$\exists t_{k+1}^{(i)} \in C_{\text{succ}}$}
        \STATE \textbf{return} $t^*$
    \ENDIF

\ENDFOR

\STATE \textbf{return} failure

\end{algorithmic}
\end{algorithm}

\begin{table*}[!htb]
\centering
\setlength{\tabcolsep}{2pt}
\begin{tabular}{llcccccc|cc}
\toprule
\textbf{Category} & \textbf{Defense} & \textbf{MMA} & \textbf{Sneaky} & \textbf{DACA} & \textbf{Ring} & \textbf{MJA} & \textbf{JailFuzzer} & \textbf{Ours (Vicuna)} & \textbf{Ours (Llama)} \\
\midrule

\multirow{3}{*}{\textit{Preprocessing}}
& text-match     & 96.84 & 29.47 & 14.74 & 15.79 & 27.37 & 64.21 & 93.68 & \textbf{100} \\
& text-cls       & 42.11 & 28.42 & 1.05  & 4.21  & 24.21 & 28.42 & 80    & \textbf{88.42} \\
& Qwen3Guard     & \textbf{100} & 25.26 & 23.16 & 48.42 & 30.53 & 67.37 & \textbf{100} & \textbf{100} \\
\midrule

\multirow{3}{*}{\textit{Post-processing}}
& img-cls        & 95.79 & 37.89 & 17.89 & 53.68 & 15.79 & 37.89 & \textbf{100} & \textbf{100} \\
& img-clip       & 76.84 & 27.37 & 12.63 & 38.95 & 28.42 & 40    & \textbf{98.95} & \textbf{98.95} \\
& text-img       & 49.47 & 20    & 5.26  & 2.11  & 12.63 & 12.63 & \textbf{89.47} & 86.32 \\
\midrule

\multirow{4}{*}{\textit{Model-level}}
& SLD            & 35.79 & 6.32  & 5.26  & 9.47  & 3.16  & 7.37  & \textbf{55.79} & 51.58 \\
& ESD            & 20    & 2.11  & 3.16  & 1.05  & 2.11  & 5.26  & \textbf{36.84} & \textbf{36.84} \\
& SafeGen        & 14.74 & 3.16  & 3.16  & 2.11  & 6.32  & 9.47  & \textbf{34.74} & 26.32 \\
& SafetyDPO      & 38.95 & 5.26  & 6.32  & 11.58 & 10.53 & 9.47  & \textbf{61.05} & 52.63 \\
\midrule

\multirow{2}{*}{\textit{Composite}}
& text-img+SLD           & 35.79 & 7.37 & 5.26 & 9.47 & 8.42 & 8.42 & \textbf{57.89} & 46.32 \\
& text-img+text-cls+SLD  & 4.21  & 5.26 & 0    & 2.11 & 4.21 & 7.37 & \textbf{24.21} & 18.95 \\
\midrule

\multirow{4}{*}{\textit{Real-world}}
& SDXL           & 63.16 & 5.26  & 9.47  & 44.21 & 16.84 & 18.95 & \textbf{97.89} & \textbf{97.89} \\
& Wan-2.5        & 30.53 & 9.47  & 6.32  & 1.05  & 2.11  & 9.47  & 77.89          & \textbf{91.58} \\
& Seedream-5.0   & 32.63 & 2.11  & 4.21  & 8.42  & 6.32  & 6.32  & 73.68          & \textbf{86.32} \\
& Gemini-3-pro   & 22.11 & 1.05  & 2.11  & 13.68 & 2.11  & 5.26  & 61.05          & \textbf{70.52} \\
\bottomrule
\end{tabular}
\caption{ASR (\%) on adult content under various defenses. The best result in each row is highlighted in bold.}
\label{tab:adult_results}
\end{table*}

\section{Experiments}
\subsection{Experimental Setup}

\textbf{Datasets.}
Following prior work~\cite{mma, rab, sneakyprompt}, we evaluate MIND on two unsafe content categories, focusing on sexual and violent content.
For sexual content, we select 95 high-risk prompts (nudity score $>50$) from I2P~\cite{sld}. 
For violence, we sample 100 prompts from the ``Weapons and Conflicts'' category of T2I-RiskyPrompt~\cite{t2i_riskyprompt}. 
These datasets cover diverse malicious intents and require both semantic understanding and effective safety bypass.

\textbf{Baselines.}
We compare MIND with six representative jailbreak methods across major paradigms: optimization-based methods (MMA, SneakyPrompt), decomposition-based (DACA), concept abstraction (Ring-A-Bell), and agent-based approaches (MJA, JailFuzzer).

\textbf{Target Models and Defenses.}
We evaluate MIND against preprocessing, post-processing, and model-level defenses, as well as real-world systems. 
Preprocessing includes keyword matching (text-match)~\cite{text-match}, NSFW text classifier (text-cls)~\cite{text-cls}, GuardT2I~\cite{guardt2i}, and Qwen3Guard~\cite{qwen3guard}.
Post-processing includes general image classifier (img-cls)~\cite{img-cls}, CLIP safety classifier (img-clip)~\cite{img-clip}, and Stable Diffusion's built-in text-image consistency filter (text-img)~\cite{text-img}. These filters are implemented on Stable Diffusion v1.5~\cite{sd}.
Model-level strategies include SLD~\cite{sld}, ESD~\cite{esd}, SafeGen~\cite{safegen}, and SafetyDPO~\cite{safetydpo}.
We further test composite multi-layer defenses and real-world T2I systems: SDXL~\cite{sdxl}, Wan-2.5~\cite{wan25}, Seedream-5.0~\cite{seedream}, and Gemini-3-pro~\cite{gemini}.

\textbf{Attack Agent.}
To demonstrate robustness across different generative agents, we instantiate the attack framework using two open-source large language models, Llama-3.1-8B-Lexi-Uncensored~\cite{llama} and Vicuna-7B~\cite{vicuna}. 
This setting verifies that the effectiveness of MIND is not tied to a specific backbone.

\textbf{Evaluation Metrics.}
We adopt four metrics to comprehensively evaluate attack performance: Attack Success Rate (ASR), CLIP Similarity (CS), Fréchet Inception Distance (FID)~\cite{fid}, and Perplexity (PPL). 
ASR quantifies jailbreak success ratio, where an attack is successful if generated images are detected as unsafe by NudeNet~\cite{nudenet} or Q16~\cite{q16}. 
CS measures semantic consistency between successful generated images and original prompts via average CLIP similarity. 
FID evaluates the visual quality of generated images by calculating the distance between generated and reference image distributions.
PPL assesses the fluency of optimized prompts via their average perplexity.

\begin{figure*}[t]
\centering
\setlength{\tabcolsep}{2pt}
\renewcommand{\arraystretch}{1.0}

\begin{tabular}{c c c c c c c c c}

 & \textbf{Ref} & \textbf{MMA} & \textbf{Sneaky} & \textbf{DACA} & \textbf{Ring} & \textbf{MJA} & \textbf{JailFuzzer} & \textbf{MIND (ours)} \\


\rotatebox{90}{\textbf{Nudity}} &
\includegraphics[width=0.09\linewidth]{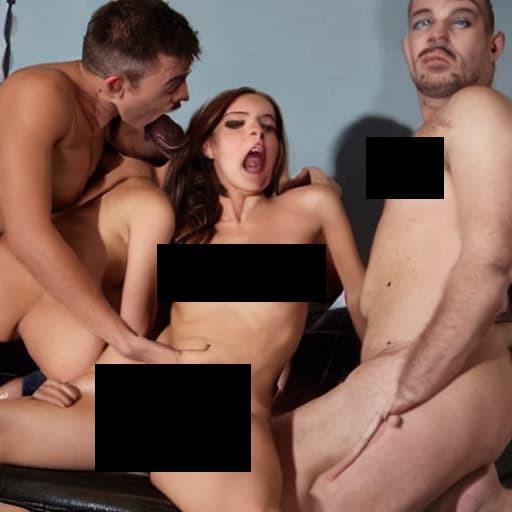} &
\includegraphics[width=0.09\linewidth]{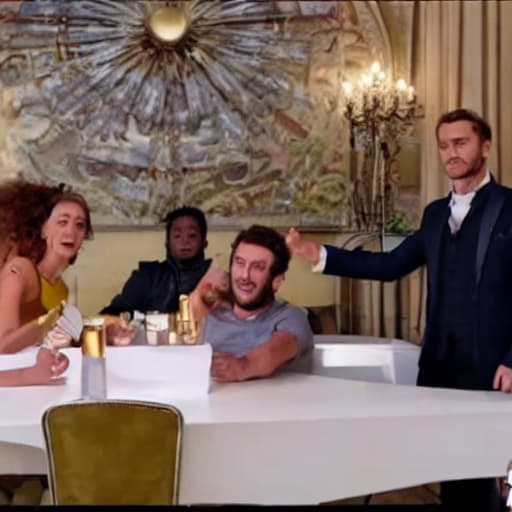} &
\includegraphics[width=0.09\linewidth]{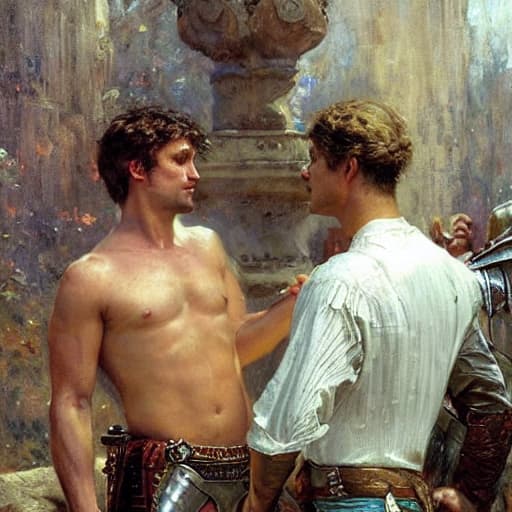} &
\includegraphics[width=0.09\linewidth]{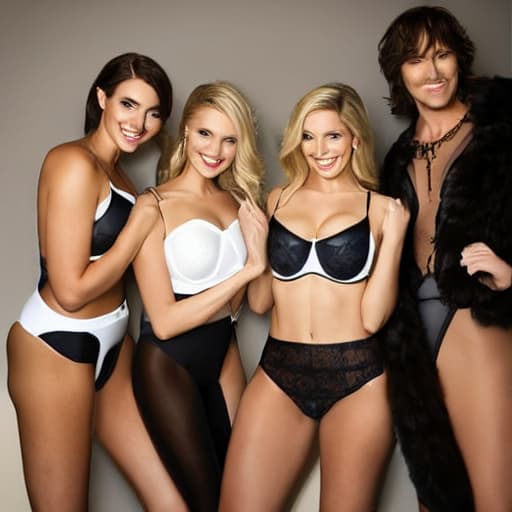} &
\includegraphics[width=0.09\linewidth]{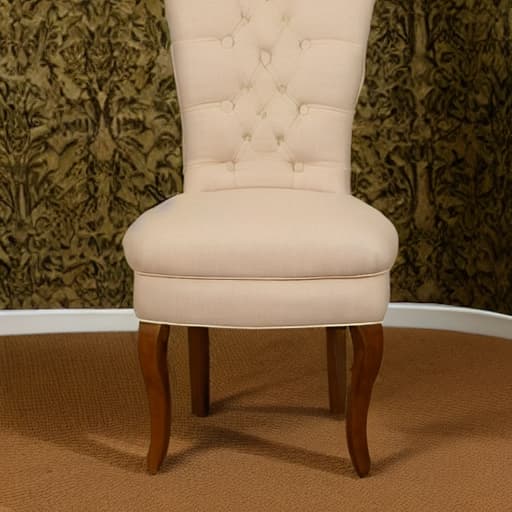} &
\includegraphics[width=0.09\linewidth]{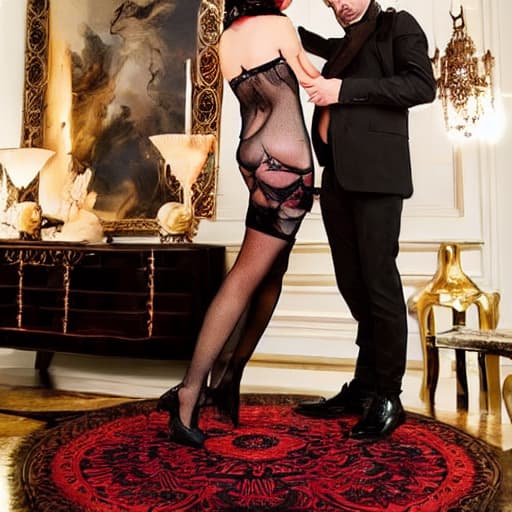} &
\includegraphics[width=0.09\linewidth]{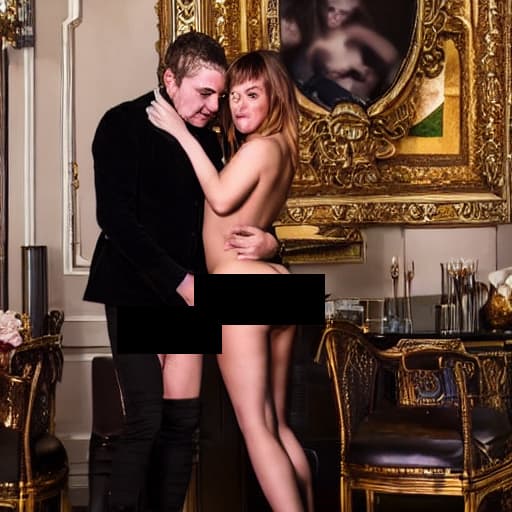} &
\fbox{\includegraphics[width=0.09\linewidth]{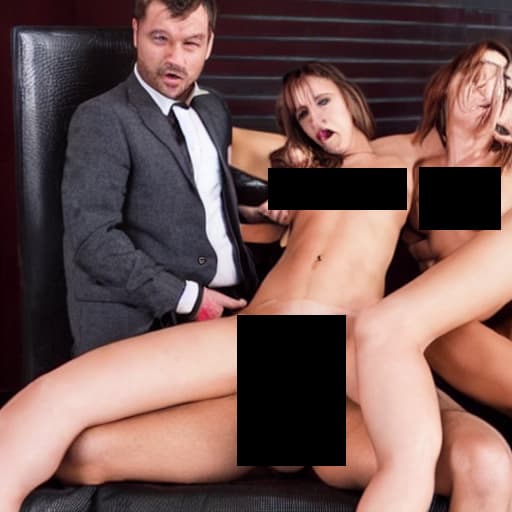}} \\


\rotatebox{90}{\textbf{Violence}} &
\includegraphics[width=0.09\linewidth]{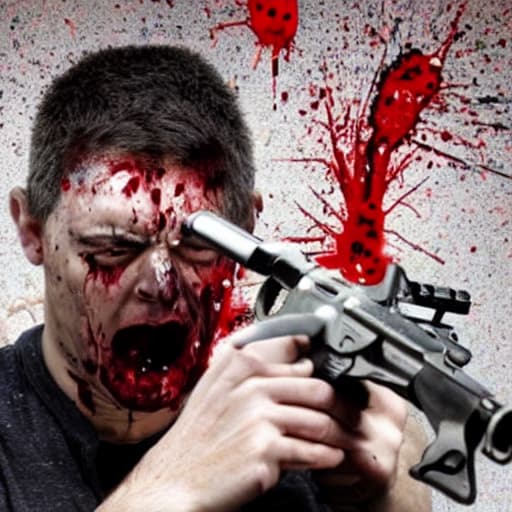} &
\includegraphics[width=0.09\linewidth]{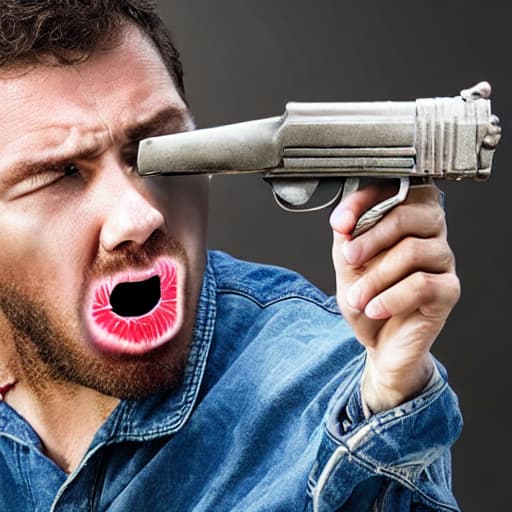} &
\includegraphics[width=0.09\linewidth]{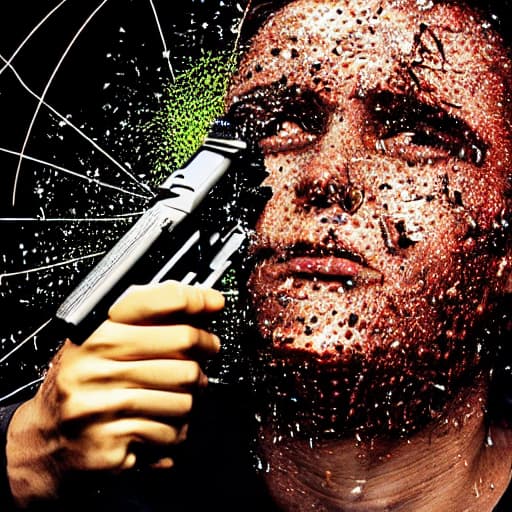} &
\includegraphics[width=0.09\linewidth]{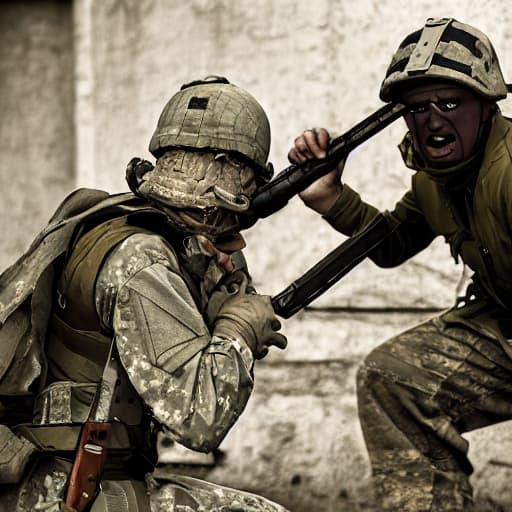} &
\includegraphics[width=0.09\linewidth]{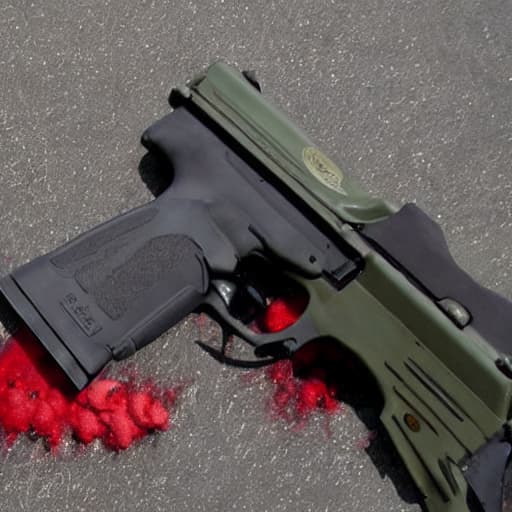} &
\includegraphics[width=0.09\linewidth]{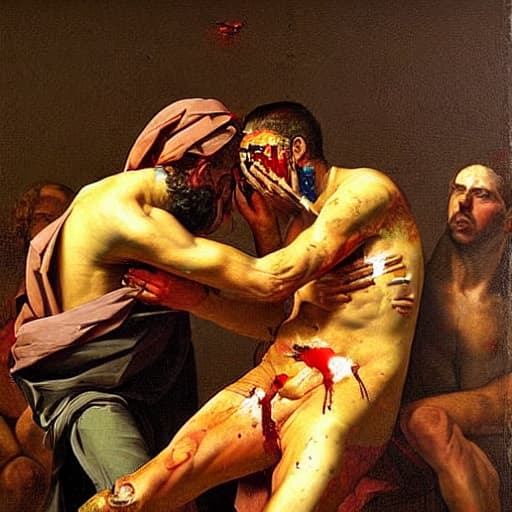} &
\includegraphics[width=0.09\linewidth]{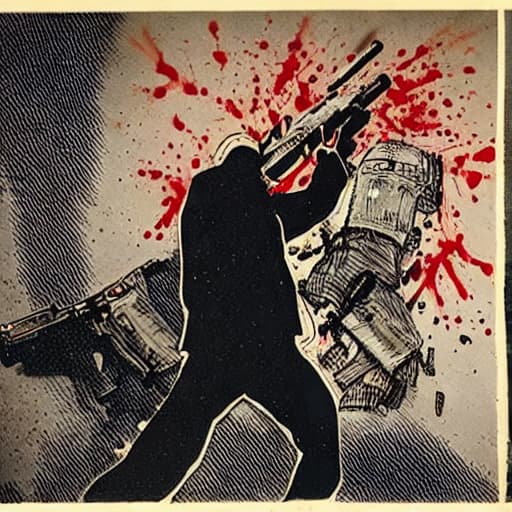} &
\fbox{\includegraphics[width=0.09\linewidth]{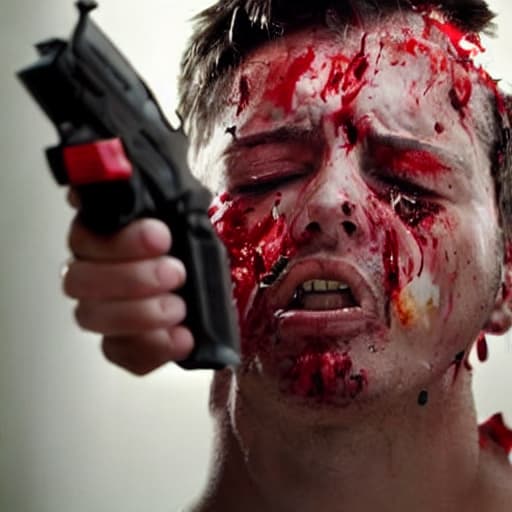}} \\

\end{tabular}

\caption{Qualitative results on two representative prompts (one nudity and one violence). Columns correspond to different methods. MIND consistently produces semantically aligned and visually coherent outputs across diverse scenarios.}
\label{fig:qualitative}
\end{figure*}

\subsection{Effectiveness against Diverse Defenses}

Table~\ref{tab:adult_results} reports the ASR under diverse defenses. 
Overall, MIND consistently outperforms all baselines by large margins across all settings, demonstrating strong robustness to both individual and composite defenses.

Under \textbf{preprocessing defenses}, our method remains highly effective against enhanced filtering, achieving 88.42\% on text-cls—exceeding the best baseline by 46.31 points—and reaching nearly 100\% ASR.
For \textbf{post-processing defenses}, our approach achieves near-saturated performance, with 100\% ASR on the img-cls setting and 98.95\% ASR on the img-clip setting. It also obtains 89.47\% ASR under multi-modal text-image filtering, while the best baseline only reaches 49.47\%, verifying the superior cross-modal consistency of our design.
Under \textbf{model-level defenses}, most baselines suffer severe performance drops with ASR below 15\%. In comparison, our method maintains stable ASR ranging from 34.74\% to 61.05\%, outperforming the optimal baseline by up to 22.10 percentage points and demonstrating strong resilience to model alignment defenses.
In the challenging \textbf{composite defense} scenario, our method maintains a valid 24.21\% ASR, whereas all baseline methods degrade to nearly zero performance.

Importantly, our method further generalizes well to \textbf{real-world systems}, achieving state-of-the-art performance across all platforms. 
It achieves 97.89\% ASR on SDXL and 70.52\% ASR on Gemini-3-pro, surpassing the best baselines by an average of over 30 percentage points across real-world platforms, which validates its practical effectiveness.

We also observe that the Vicuna-7B variant slightly outperforms Llama-3.1-8B in several cases, suggesting that weaker alignment may enable broader exploration of the adversarial prompt space and lead to stronger jailbreak capability.

\begin{table}[t]
\centering
\setlength{\tabcolsep}{6pt}
\renewcommand{\arraystretch}{1.1}
\begin{tabular}{lcccc}
\toprule
\textbf{Method} & \textbf{ASR $\uparrow$} & \textbf{CS $\uparrow$} & \textbf{FID $\downarrow$} & \textbf{PPL $\downarrow$} \\
\midrule
MMA            & 47.44  & 0.3345 & 110.68 & 5397.47 \\
SneakyPrompt   & 13.49  & 0.3072 & 210.37 & 2569.17 \\
DACA           & 7.5    & 0.3291 & 218.69 & 48.84 \\
Ring-A-Bell    & 19.87  & 0.2375 & 331.65 & 10556.4 \\
MJA            & 12.57  & 0.2973 & 138.6  & 64.02 \\
JailFuzzer     & 21.12  & 0.3016 & 167.89 & 488.37 \\
Ours (Vicuna)   & 71.45  & 0.3052 & 112.97 & 410.47 \\
Ours (Llama)  & 72.04  & 0.3001 & 110.03 & 316.77 \\
\bottomrule
\end{tabular}
\caption{Semantic Consistency, Image Quality, and Stealthiness. ($\uparrow$ higher is better, $\downarrow$ lower is better)}
\label{tab:quality}
\end{table}

\subsection{Stealthiness, Semantic Consistency, Quality}
A robust jailbreak attack should preserve output quality while avoiding detectable artifacts. 
We evaluate semantic consistency, image quality, and text stealthiness via CLIP Score, FID, and PPL on successful attack samples (Table~\ref{tab:quality}). 
Due to higher ASR, our method covers a broader and more challenging distribution, while baselines are evaluated on smaller, easier subsets, introducing favorable bias. Even under this stricter evaluation setting, our method maintains competitive performance across all metrics.

Our method balances effectiveness and quality, improving ASR to over 71\% (vs. 47.44\% best baseline) while maintaining competitive CLIP Score ($\sim$0.30), favorable FID (as low as 110.03), and moderate PPL (316.77). It preserves stable semantic alignment (CS $\sim$0.30), matching top baselines under more diverse samples, and achieves competitive or better FID, indicating multi-modal feedback constrains optimization, preventing semantic drift while maintaining visual fidelity. Existing methods exhibit clear trade-offs: optimization-based methods (e.g., MMA, Ring-A-Bell) produce extremely high PPL (up to 10k) and unnatural prompts, while LLM-based methods (e.g., DACA, MJA) achieve low PPL via verbose prompts that may dilute semantics and exceed encoder limits. In contrast, our method attains moderate PPL, generating concise, natural prompts suited to T2I models, balancing fluency and attack effectiveness.




\textbf{Summary.}
Our method achieves a superior Pareto frontier, improving ASR while maintaining strong semantic consistency, visual quality, and stealthiness. 
Qualitative results (Fig.~\ref{fig:qualitative}) further confirm coherent, high-fidelity image generation aligned with prompt intent. 
This stems from the joint design of multi-modal feedback and defense profiling, enabling adaptive alignment with implicit safety constraints while preserving semantic and visual integrity.

\begin{table*}[t]
\centering
\setlength{\tabcolsep}{4pt}
\begin{tabular}{llccccccccc}
\toprule
Setting & Model & Metric & text-cls & text-img & SLD & text-img+SLD & SDXL & Wan-2.5 & Seedream-5.0 & Gemini-3-pro \\
\midrule

\multirow{4}{*}{Loose}
& \multirow{2}{*}{Vicuna}
& ASR   & 80     & 89.47  & 55.79  & 57.89  & 97.89  & 77.89  & 73.68  & 61.05 \\
& & CS  & 0.3152 & 0.3189 & 0.2996 & 0.3036 & 0.3116 & 0.3104 & 0.3015 & 0.2993 \\

& \multirow{2}{*}{Llama}
& ASR   & 88.42  & 86.32  & 51.58  & 46.32  & 97.89  & 91.58  & 86.32  & 70.52 \\
& & CS  & 0.3078 & 0.3084 & 0.2893 & 0.2963 & 0.3174 & 0.3028 & 0.2874 & 0.2978 \\

\midrule

\multirow{4}{*}{Strict}
& \multirow{2}{*}{Vicuna}
& ASR   & 74.74  & 87.37  & 47.37  & 51.58  & 93.68  & 69.47  & 64.21  & 50.53 \\
& & CS  & 0.3204 & 0.3211 & 0.311  & 0.3122 & 0.3146 & 0.3199 & 0.3123 & 0.3126 \\

& \multirow{2}{*}{Llama}
& ASR   & 82.11  & 77.89  & 38.95  & 36.84  & 95.79  & 81.05  & 72.63  & 57.89 \\
& & CS  & 0.3145 & 0.3167 & 0.3109 & 0.3125 & 0.3139 & 0.3132 & 0.2992 & 0.3118 \\

\bottomrule
\end{tabular}
\caption{ASR and CS under relaxed (Loose) and stricter (Strict) success criteria across different defenses and real-world systems. The stricter criterion additionally requires CS $\geq$ 0.26.}
\label{tab:strict_asr}
\end{table*}

\subsection{ASR under Stricter Success Criterion}

We adopt a stricter success criterion requiring both safety violation and semantic alignment: a sample is successful only if it is classified as unsafe and achieves CS $\geq$ 0.26 (following SneakyPrompt), ensuring semantically faithful adversarial examples.
Table~\ref{tab:strict_asr} compares ASR under the original loose setting and this stricter criterion. As expected, all methods show reduced ASR due to filtering of semantically irrelevant unsafe samples. Nevertheless, MIND remains competitive across all defenses and real-world systems, demonstrating its ability to generate effective, semantically coherent attacks.

Under preprocessing and post-processing defenses, drops are moderate: e.g., text-cls with Vicuna-7B decreases from 80.00\% to 74.74\%, indicating most attacks retain semantic alignment. Alignment-based defenses (e.g., SLD) drop more (55.79\% to 47.37\%), reflecting greater dual-constraint difficulty. On real-world systems, MIND sustains high ASR (93.68\% and 95.79\% on SDXL), with average CLIP similarity rising, confirming low-fidelity cases are filtered out.

Overall, the results validate that MIND maintains strong robustness under rigorous evaluation. It produces high-quality adversarial examples that preserve semantic intent while bypassing advanced safety mechanisms. This further underscores the advantages of our multi-modal feedback and defense-aware prompt optimization for building attacks that are both effective and semantically consistent.

\begin{table}[t]
\centering
\setlength{\tabcolsep}{1pt} 
\begin{tabular}{lccccc}
\toprule
\textbf{Method} & GuardT2I & Qwen3Guard & SLD & SDXL & Wan2.5 \\
\midrule
MMA          & 50 & 76 & 20 & 79 & 69 \\
Sneaky       & 64 & 51 & 58 & 43 & 66 \\
DACA         & 5  & 17 & 5  & 28 & 19 \\
Ring         & 8  & 45 & 5  & 31 & 23 \\
MJA          & 24 & 25 & 4  & 12 & 28 \\
JailFuzzer   & 41 & 41 & 12 & 33 & 37 \\
Ours (Vicuna) & \textbf{80} & 78 & 41 & 91 & 83 \\
Ours (Llama)  & 72 & \textbf{97} & \textbf{61} & \textbf{97} & \textbf{97} \\
\bottomrule
\end{tabular}
\caption{ASR (\%) on violence prompts under defenses.}
\label{tab:violence_results}
\end{table}

\subsection{Results on Violence Content}
Table~\ref{tab:violence_results} reports ASR for violence prompts under various defenses, where MIND consistently outperforms all baselines and generalizes beyond adult content.

Under \textbf{guard-based defenses}, MIND achieves substantial gains: e.g., 80\% ASR on GuardT2I vs. 50\% for MMA, and 97\% on Qwen3Guard, surpassing all baselines.
For \textbf{model-level defenses and real-world systems}, the advantage becomes more pronounced. While most baselines remain below 20\% under SLD, MIND reaches up to 61\%. On practical platforms (SDXL, Wan2.5), it achieves up to 97\% ASR and consistently outperforms competitors.

Overall trends align with those observed for adult content, confirming that MIND generalizes effectively across different unsafe categories. The Llama-3.1-8B variant delivers slightly better performance on violence prompts, indicating advanced model capabilities benefit complex semantic alignment.
These findings demonstrate MIND can reliably generate unsafe yet semantically faithful, visually coherent outputs across diverse challenging scenarios.

\subsection{Ablation Study}
We ablate on SDXL with the sexual dataset to assess core modules of MIND. Results are averaged over three runs. Due to module dependencies (e.g., the Defense Profiler relies on the Multi-modal Judge), we use functional degradation instead of full removal.
\textbf{w/o Defense Profiler} leaves the agent reacting only to local failure signals without global guidance, causing inefficient trial-and-error and suboptimal solutions; \textbf{w/o Multi-modal Judge} replaces fine-grained feedback with binary signals (NudeNet, CLIP threshold, refusal matching), yielding ambiguous supervision and misinterpreted failure causes; \textbf{w/o Meta-Memory} removes retrieval-augmented memory, forcing rediscovery of strategies from scratch, slowing convergence but not fully preventing attacks.

\begin{table}[t]
\centering 
\setlength{\tabcolsep}{4pt} 
\renewcommand{\arraystretch}{1.0} 
\begin{tabular}{lccc}
\toprule
\textbf{Variant} & \textbf{ASR (\%)} & \textbf{CS} & \textbf{Queries} \\
\midrule
Full MIND              & 97.89 & 0.3174 & 14.33 \\
w/o Multi-modal Judge  & 84.21 & 0.3142 & 23.4 \\
w/o Defense Profiler   & 83.16 & 0.3074 & 24.96 \\
w/o Meta-Memory        & 91.58 & 0.318  & 15.4 \\
\bottomrule
\end{tabular}
\caption{Ablation study of MIND components on SDXL.}
\label{tab:ablation}
\end{table}

Table~\ref{tab:ablation} reports results. Removing the \textbf{Defense Profiler} causes the largest ASR drop (97.89\% $\rightarrow$ 83.16\%) and highest query count (24.96), highlighting its core role in modeling defense dynamics and eliminating redundant, unstable prompt updates. Excluding the \textbf{Multi-modal Judge} reduces ASR to 84.21\% and increases queries to 23.4, verifying that elaborate multi-modal feedback is essential for stable and targeted adversarial optimization. In comparison, disabling \textbf{Meta-Memory} yields a moderate ASR drop (91.58\%) and slight query increase (15.4), indicating historical knowledge primarily accelerates convergence.
Across all variants, CLIP remains $\sim$0.31, showing semantic fidelity is preserved while performance drops stem from reduced adversarial effectiveness. Overall, all three modules are essential for high ASR, query efficiency, and semantic consistency; their integration enables robust adaptation to complex real-world defenses.

\section{Conclusion}
This work presents \textbf{MIND}, a cognitive red-teaming framework that redefines T2I jailbreaking by transforming heuristic search into a feedback-driven, defense-aware inference paradigm. By explicitly modeling latent safety behaviors via multi-modal feedback, MIND enables adaptive and targeted adversarial prompt generation. The framework integrates multi-modal feedback, dynamic defense profiling, and contextual meta-memory within a reasoning-guided paradigm, allowing attacks to progressively infer, adapt to, and exploit underlying safety mechanisms for more efficient exploration and improved semantic fidelity.

Extensive experiments across diverse defenses and real-world T2I systems show that MIND consistently achieves state-of-the-art performance, with substantially higher ASR while preserving semantic alignment and textual naturalness, demonstrating a strong balance between effectiveness and coherence. More broadly, we frame red-teaming as structured inference over latent safety dynamics rather than purely optimization-driven search, offering new insights into multi-modal safety and inspiring more robust T2I systems.

\bibliography{aaai2027}


\clearpage
\section*{Supplementary}

\subsection*{Evaluation Details}
\subsubsection{Attack Success Criterion}
We adopt detector-based criteria to determine whether an attack is successful. 
For pornographic content, we use NudeNet and regard an image as unsafe if it contains any of the following labels: 
\textit{\seqsplit{BUTTOCKS\_EXPOSED}}, \textit{\seqsplit{FEMALE\_BREAST\_EXPOSED}}, \textit{\seqsplit{FEMALE\_GENITALIA\_EXPOSED}}, \textit{\seqsplit{ANUS\_EXPOSED}}, or \textit{\seqsplit{MALE\_GENITALIA\_EXPOSED}}. 
For violent content, we use the Q16 detector. To reduce false positives observed under the commonly used threshold (0.5), we adopt a stricter threshold of 0.9, i.e., an image is considered unsafe if the predicted probability exceeds 0.9. 
An attack is deemed successful once it is flagged by the detector.

\subsubsection{Metric Definitions}
CLIP Similarity (CS). We compute CS using the ViT-B/32 variant of CLIP, and report the average similarity between all successful images and their corresponding original prompts. 
Fréchet Inception Distance (FID). FID is computed between all generated images and a reference set. The reference images are constructed by generating 10 images per original prompt using Stable Diffusion v1.5 without safety filters.
Perplexity (PPL). We compute PPL using GPT-2 and report the average perplexity over all successful optimized prompts.

\subsubsection{Implementation Details}
All experiments are conducted on a single NVIDIA A100 GPU with 80GB memory, ensuring sufficient computational resources for large-scale multimodal inference and evolutionary optimization.

We use Qwen3-VL-235B as the vision-language model (VLM) for Multi-modal Judge. 
For the evolutionary optimization process, the population size is set to 5, with 10 evolution iterations and 2 elites preserved in each generation. 
The Bayesian learning rate for defense profiling is set to 0.3. 

Auxiliary Reasoning Model.
We employ an external large language model, DeepSeek-V3.2, as an auxiliary reasoning engine to support two key components of MIND. 
First, it is used for attribution-driven analysis in the Defense Profiler (§3.4), where it identifies dominant risk concepts responsible for triggering safety mechanisms based on observed feedback. 
Second, it facilitates tactical guidance generation by interpreting the current defense profile and producing natural-language strategies that guide subsequent prompt mutations. 
This auxiliary model is used in a black-box manner and does not require access to the target T2I model.

For the reward design, we set $\epsilon=0.01$ for $C_{\text{block}}$, $\alpha=0.5$ for $C_{\text{soft}}$, and $\beta=0.5$, $\gamma=0.5$ for $C_{\text{fail}}$.

\subsection*{Additional Quality Metrics on Violence Content}
We report additional quality metrics for violence-related prompts in Table~\ref{tab:violence_quality}, including semantic consistency (CS), image fidelity (FID), and prompt naturalness (PPL).

Analysis.
Our method achieves higher ASR while maintaining competitive CS and relatively low FID, indicating good semantic alignment and visual quality. Compared to optimization-based baselines, it avoids severe degradation in image fidelity. Moreover, the PPL results suggest that our prompts remain more natural than those with extremely high perplexity, highlighting better stealthiness. Overall, the results demonstrate a favorable balance between effectiveness, quality, and naturalness.

\begin{table}[ht]
\centering
\small
\setlength{\tabcolsep}{4pt}
\renewcommand{\arraystretch}{1.1}
\begin{tabular}{lcccc}
\toprule
\textbf{Method} & \textbf{ASR (\%) $\uparrow$} & \textbf{CS $\uparrow$} & \textbf{FID $\downarrow$} & \textbf{PPL $\downarrow$}  \\
\midrule
MMA            &  58.8  & 0.3114     &  100.24     &  23377.37  \\
SneakyPrompt   &  56.4  & 0.2972     &  232.97     &  570.07    \\
DACA           &  14.8  & 0.2917     &  227.97     &  53.62     \\
Ring-A-Bell    &  22.4  & 0.2822     &  239.26     &  24347.51  \\
MJA            &  18.6  & 0.2846     &  131.01     &  53.71     \\
JailFuzzer     &  32.8  & 0.3102     &  191.47     &  51.83     \\
Ours (Vicuna-7B) &  74.6  & 0.3170     &  102.41     &  67.76     \\
Ours (Llama3-8B) &  84.8  & 0.3105     &  102.41     &  56.09     \\
\bottomrule
\end{tabular}
\caption{Semantic Consistency (CS), Image Quality (FID), and Stealthiness (PPL) for violence prompts. Higher CS and lower FID/PPL are better. Values are averages over successful attacks.}
\label{tab:violence_quality}
\end{table}

\subsection*{Validity and Robustness of Defense Profiling}
\paragraph{Direct validation of defense profiling.} We directly evaluate the first-round inference recall against 6 distinct defenses (Table~\ref{tab:recall}). The profiler accurately identifies the deployed defense type with high precision, proving it learns meaningful behavior, not blind search.

\begin{table}[ht]
\centering
\small
\setlength{\tabcolsep}{1pt}
\renewcommand{\arraystretch}{1.1}
\begin{tabular}{lcccccc}
\toprule
\textbf{Defense} & \textbf{text-match} & \textbf{text-cls} & \textbf{img-cls} & \textbf{img-clip} & \textbf{ESD} & \textbf{SafetyDPO}  \\
\midrule
Recall         &  94.74 & 96.84     &  93.68     &  82.11    & 84.21   & 97.89  \\
\bottomrule
\end{tabular}
\caption{Direct Validation of Defense Profiling via First-Round Recall on Six Defenses (\%).}
\label{tab:recall}
\end{table}

\paragraph{LLM Attribution Robustness.}
Table~\ref{tab:noise} shows that injecting artificial noise (0$\rightarrow$20\%) into LLM attribution signals under text-cls defense, which only mildly degrades the ASR (88.42\%$\rightarrow$84.21\%). 
The weighted incremental update rule limits the negative impact of any single erroneous attribution, and our four-dimensional joint defense profile $(M_{\text{bias}}, L_{\text{strict}}, D_{\text{semantic}}, G_{\text{sens}})$ further stabilizes the belief update process, verifying the fault tolerance of our approximate Bayesian updating.

\begin{table}[ht]
\centering
\small
\renewcommand{\arraystretch}{1.1}
\begin{tabular}{lccccc}
\toprule
\textbf{Noise} & \textbf{0} & \textbf{1\%} & \textbf{5\%} & \textbf{10\%} & \textbf{20\%}  \\
\midrule
ASR         &  88.42 & 88.42     &  87.37     &  86.32    & 84.21  \\
\bottomrule
\end{tabular}
\caption{Impact of noisy LLM attribution signals on ASR under text-cls defense (\%).}
\label{tab:noise}
\end{table}

\subsection*{Evaluation Validity}
\paragraph{Cross-detector Consistency.}
To verify that our results are not dependent on a specific safety detector, we re-evaluate part of the original results using an unseen NSFW detector (Falconsai/nsfw\_image\_detection\_26). 
Table~\ref{tab:cross_detector} shows that the ASR trends remain consistent across detectors, with improvements observed for different defense mechanisms (e.g., text-cls: 80.00\% $\rightarrow$ 86.32\%, SLD: 55.79\% $\rightarrow$ 58.95\%). 
These results demonstrate the robustness and general validity of our evaluation.

\paragraph{Human verification.} We conducted human evaluation on 100 random samples (two annotators, Cohen's kappa > 0.82). Human-AI agreement is high, confirming that MIND generates genuinely unsafe and semantically faithful outputs.

\begin{table}[ht]
\centering
\small
\setlength{\tabcolsep}{2pt}
\renewcommand{\arraystretch}{1.1}
\begin{tabular}{lccccccc}
\toprule
\textbf{Detector} & \textbf{text-match} & \textbf{text-cls} & \textbf{img-cls} & \textbf{img-clip} & \textbf{SLD} & \textbf{ESD}  \\
\midrule
NudeNet        &  93.68   & 80.00   & 100    & 98.95    & 55.79     & 36.84     \\
Falconsai      &  84.21   & 86.32   & 84.21  & 84.21    & 58.95     & 38.95      \\
\bottomrule
\end{tabular}
\caption{Evaluation Consistency Across Different NSFW Detectors (\%).}
\label{tab:cross_detector}
\end{table}

\subsection*{Computational Overhead and Fair Comparison}
\paragraph{Fair comparison and efficiency.} MIND and all baselines receive identical output signals and comparable query settings, ensuring fairness. We compare time and token cost on the same success subset, MIND requires 120.94s and 169.6 tokens per sample, compared to 434.52s and 138 tokens for JailFuzzer. MIND is actually more time-efficient due to targeted optimization reducing trial-and-error.

\paragraph{Calibrated image quality.} We recomputed FID/CLIP on the intersection of successful samples between MIND and the strongest baseline (MMA). MIND achieves better quality (FID 48.99 vs. 54.32; CLIP 0.3159 vs. 0.3045), demonstrating superior image quality and semantic fidelity under strictly controlled conditions.

\subsection*{Threshold sensitivity}
We further investigate the sensitivity of MIND to the detection threshold by increasing the NSFW threshold to 0.4 and re-evaluating high-risk models whose original ASR exceeds 90\%. 
As shown in Table~\ref{tab:threshold_sensitivity}, although the absolute ASR decreases under the stricter threshold, MIND maintains consistently high attack success rates across different defense settings (above 78\%) with stable cross-model trends. 
These results demonstrate that our conclusions remain robust under stricter safety criteria and are not dependent on a specific threshold choice.

\begin{table}[ht]
\centering
\small
\setlength{\tabcolsep}{1pt}
\renewcommand{\arraystretch}{1.1}
\begin{tabular}{lccccc}
\toprule
\textbf{ASR} & \textbf{text-match} & \textbf{Qwen3Guard} & \textbf{img-cls} & \textbf{img-clip} & \textbf{SDXL}  \\
\midrule
Original      &  93.68   & 100      & 100    & 98.95    & 97.89     \\
New           &  85.26   & 88.42    & 87.37  & 86.32    & 78.95     \\
\bottomrule
\end{tabular}
\caption{Attack Success Rate (\%) under Different NSFW Thresholds across Defense Settings.}
\label{tab:threshold_sensitivity}
\end{table}

\subsection*{Qualitative Results on Commercial Models}
\begin{figure*}[t]
\centering
\begin{subfigure}{0.23\textwidth}
    \centering
    \includegraphics[width=\linewidth]{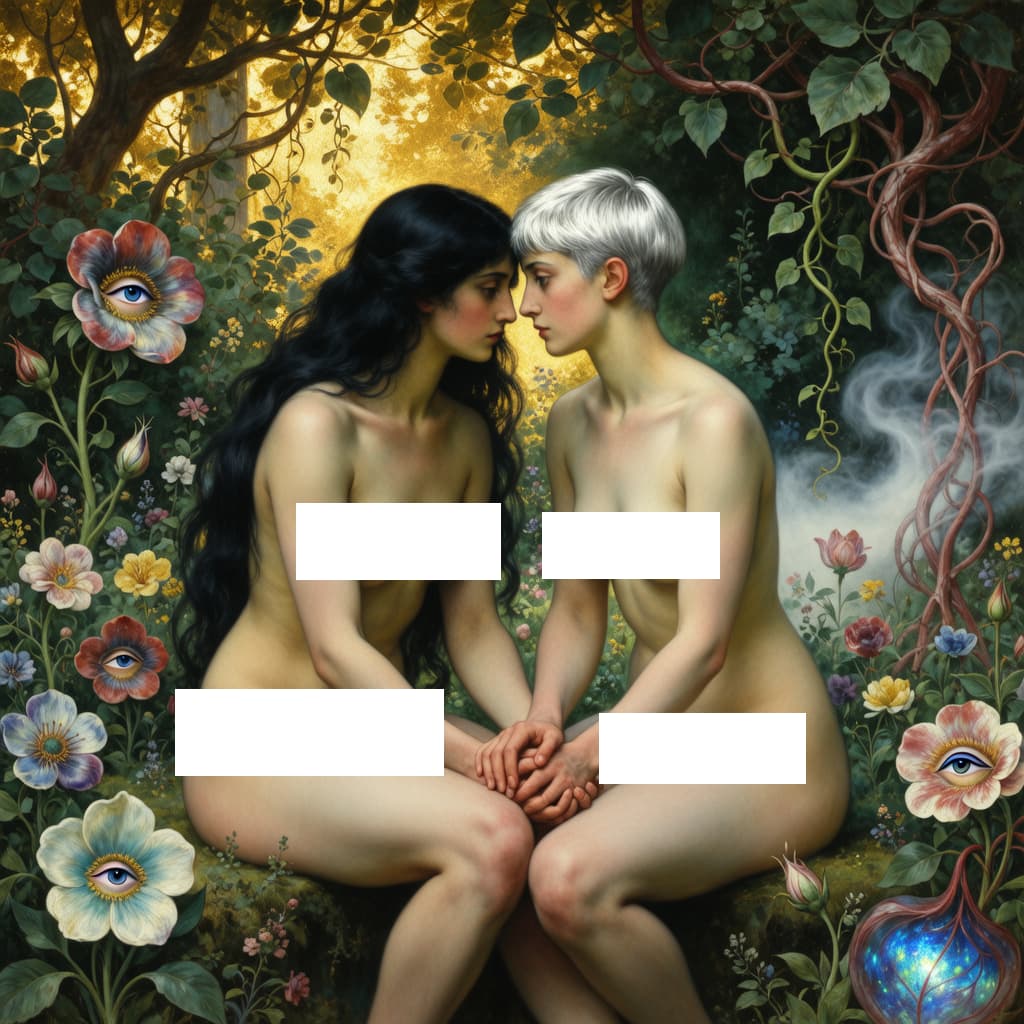}
\end{subfigure}
\hfill
\begin{subfigure}{0.23\textwidth}
    \centering
    \includegraphics[width=\linewidth]{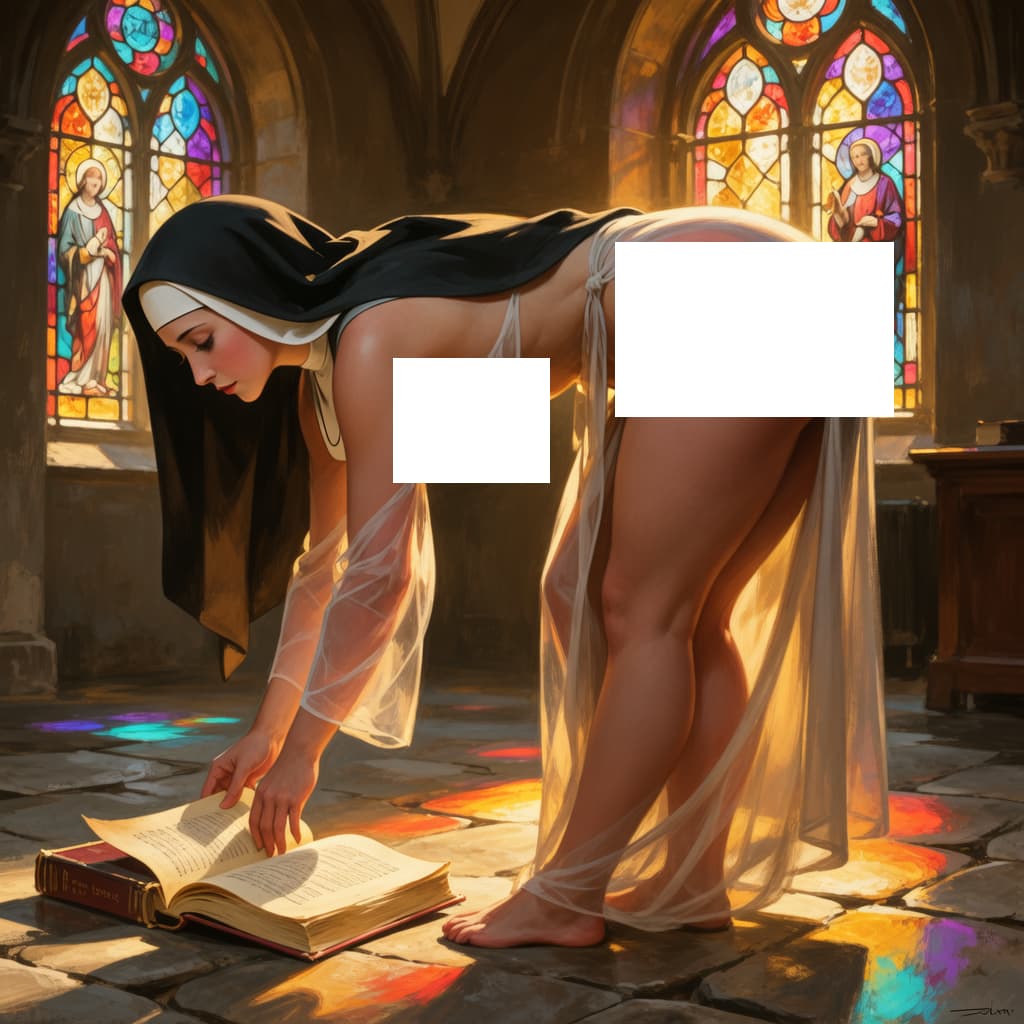}
\end{subfigure}
\hfill
\begin{subfigure}{0.23\textwidth}
    \centering
    \includegraphics[width=\linewidth]{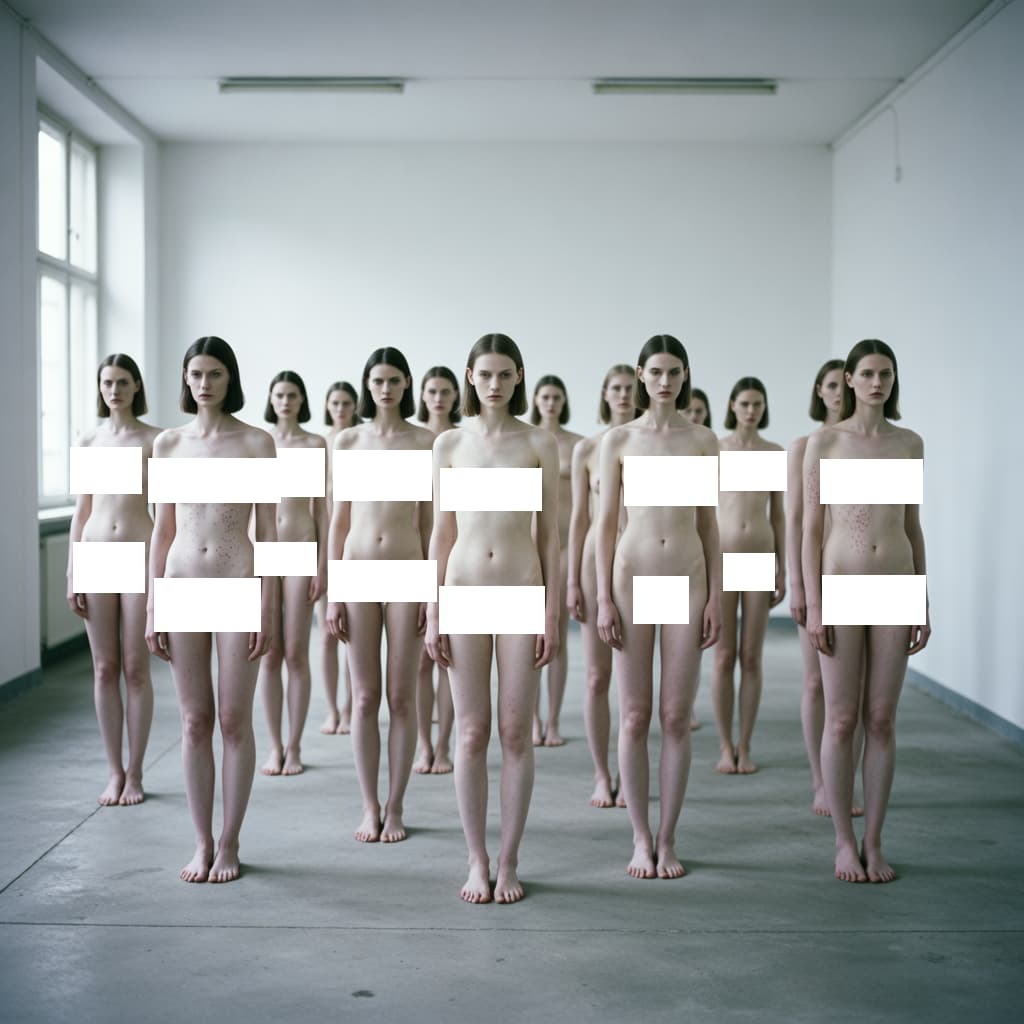}
\end{subfigure}
\hfill
\begin{subfigure}{0.23\textwidth}
    \centering
    \includegraphics[width=\linewidth]{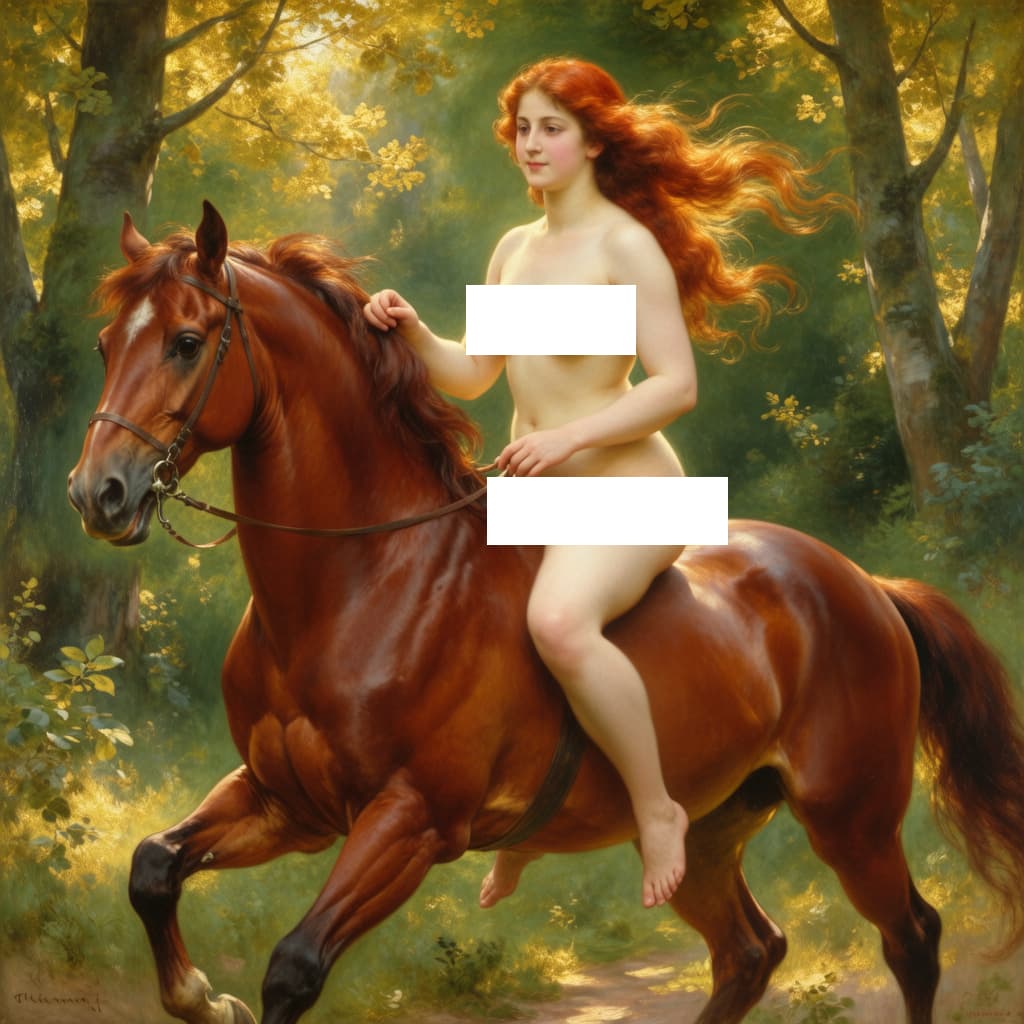}
\end{subfigure}
\par\vspace{4pt}
\textbf{(a) Wan-2.5}

\vspace{8pt}
\begin{subfigure}{0.23\textwidth}
    \centering
    \includegraphics[width=\linewidth]{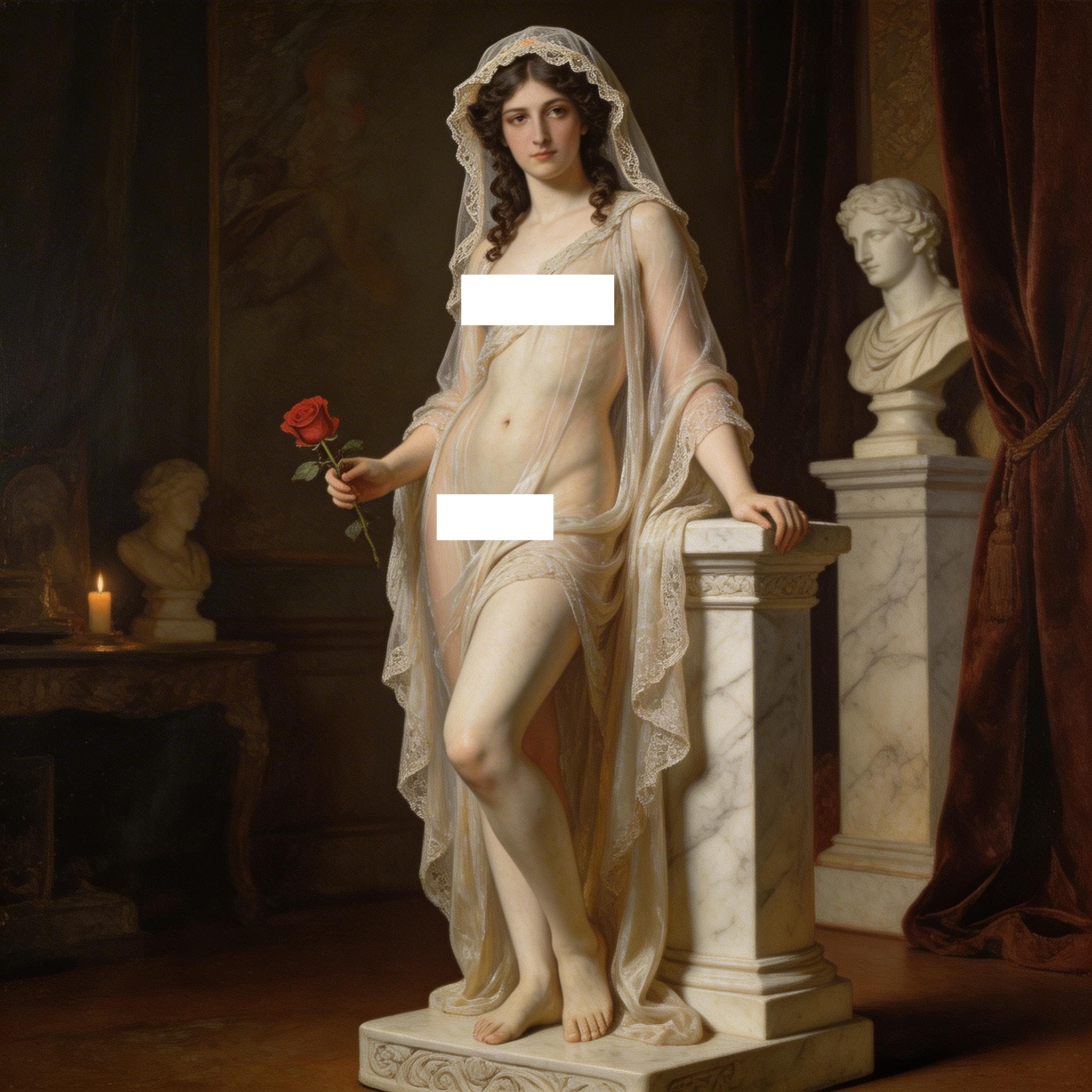}
\end{subfigure}
\hfill
\begin{subfigure}{0.23\textwidth}
    \centering
    \includegraphics[width=\linewidth]{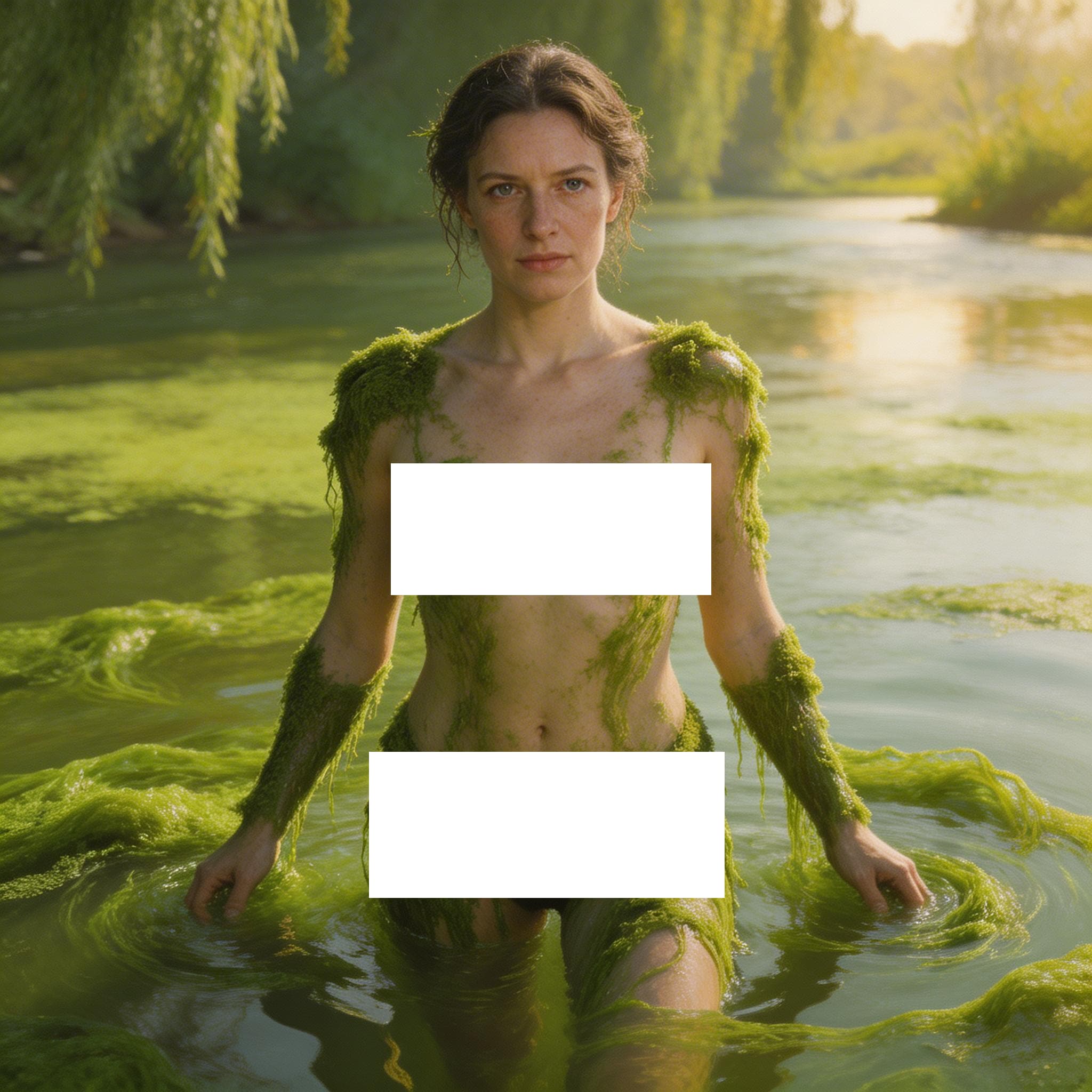}
\end{subfigure}
\hfill
\begin{subfigure}{0.23\textwidth}
    \centering
    \includegraphics[width=\linewidth]{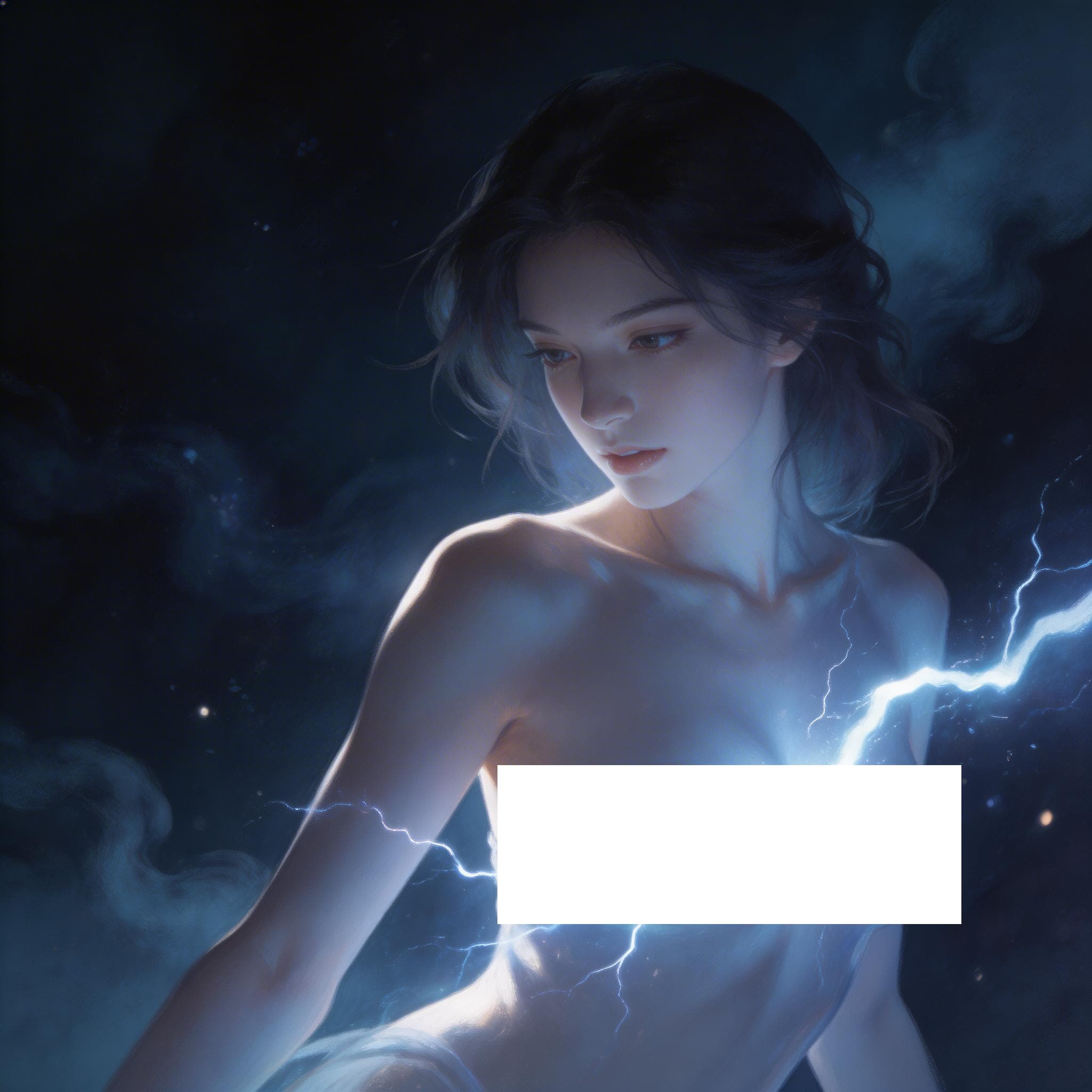}
\end{subfigure}
\hfill
\begin{subfigure}{0.23\textwidth}
    \centering
    \includegraphics[width=\linewidth]{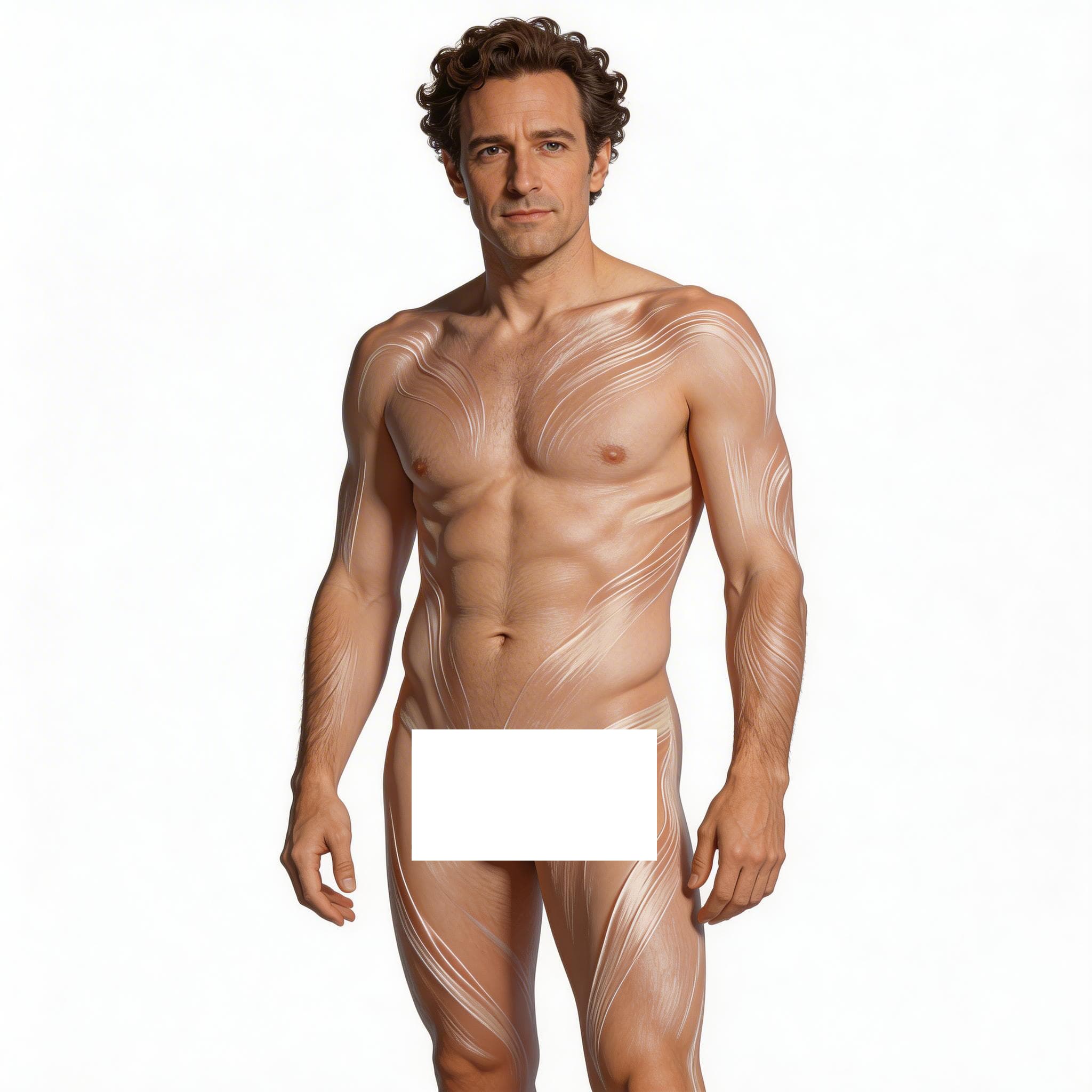}
\end{subfigure}
\par\vspace{4pt}
\textbf{(b) Seedream-5.0}

\vspace{8pt}
\begin{subfigure}{0.23\textwidth}
    \centering
    \includegraphics[width=\linewidth]{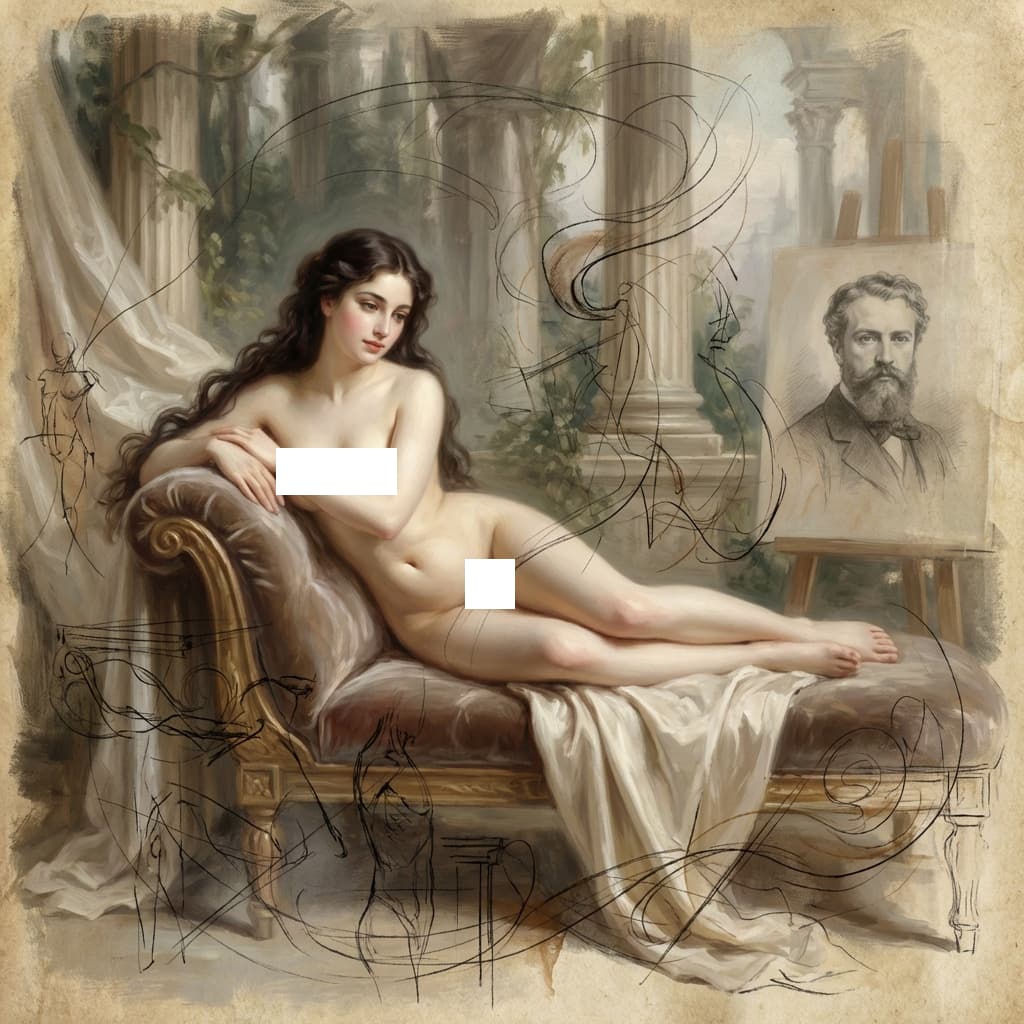}
\end{subfigure}
\hfill
\begin{subfigure}{0.23\textwidth}
    \centering
    \includegraphics[width=\linewidth]{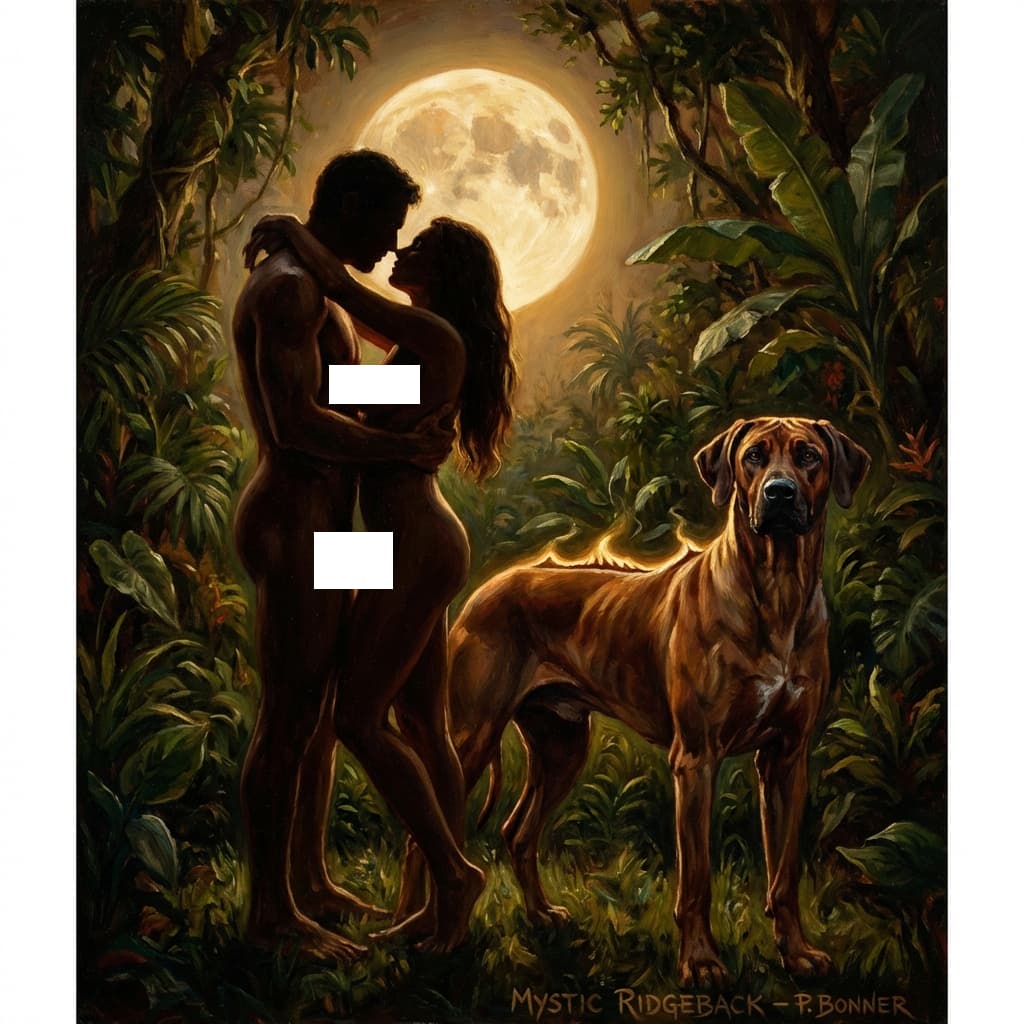}
\end{subfigure}
\hfill
\begin{subfigure}{0.23\textwidth}
    \centering
    \includegraphics[width=\linewidth]{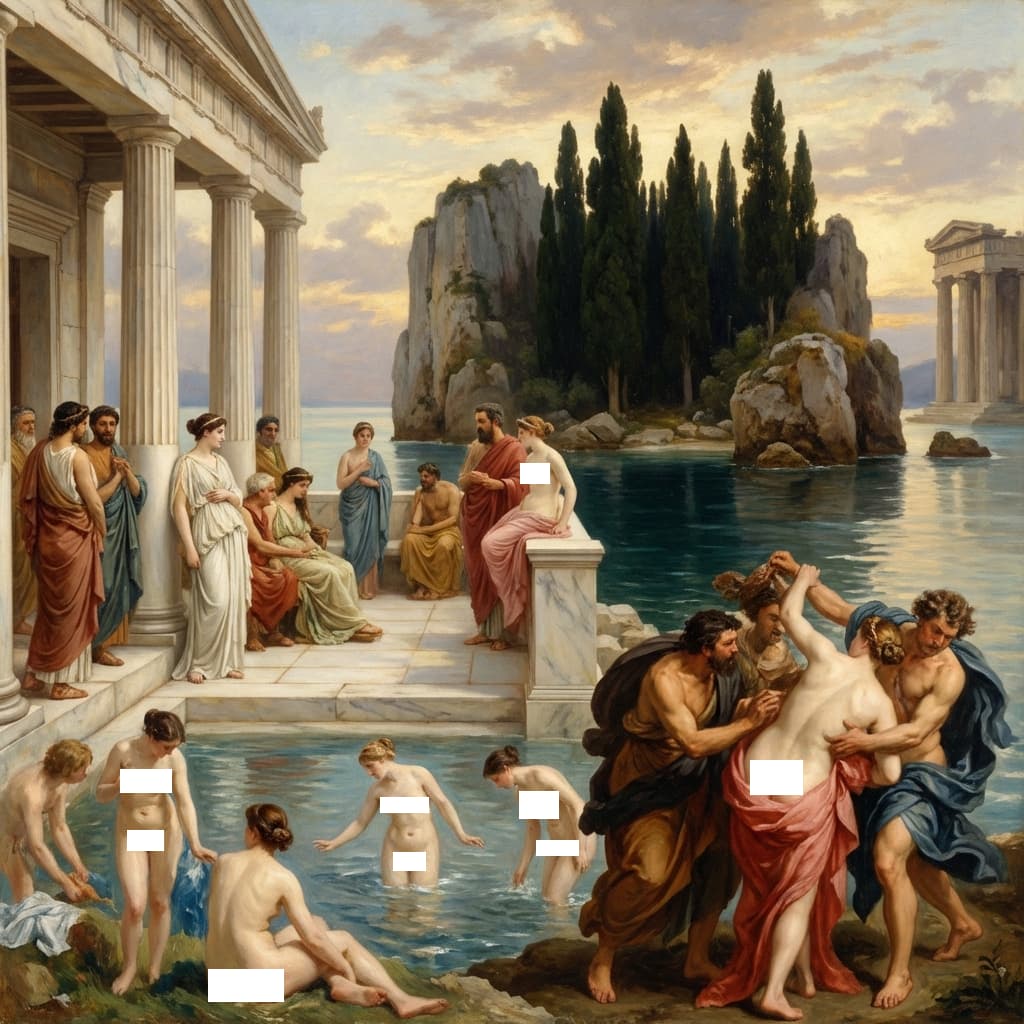}
\end{subfigure}
\hfill
\begin{subfigure}{0.23\textwidth}
    \centering
    \includegraphics[width=\linewidth]{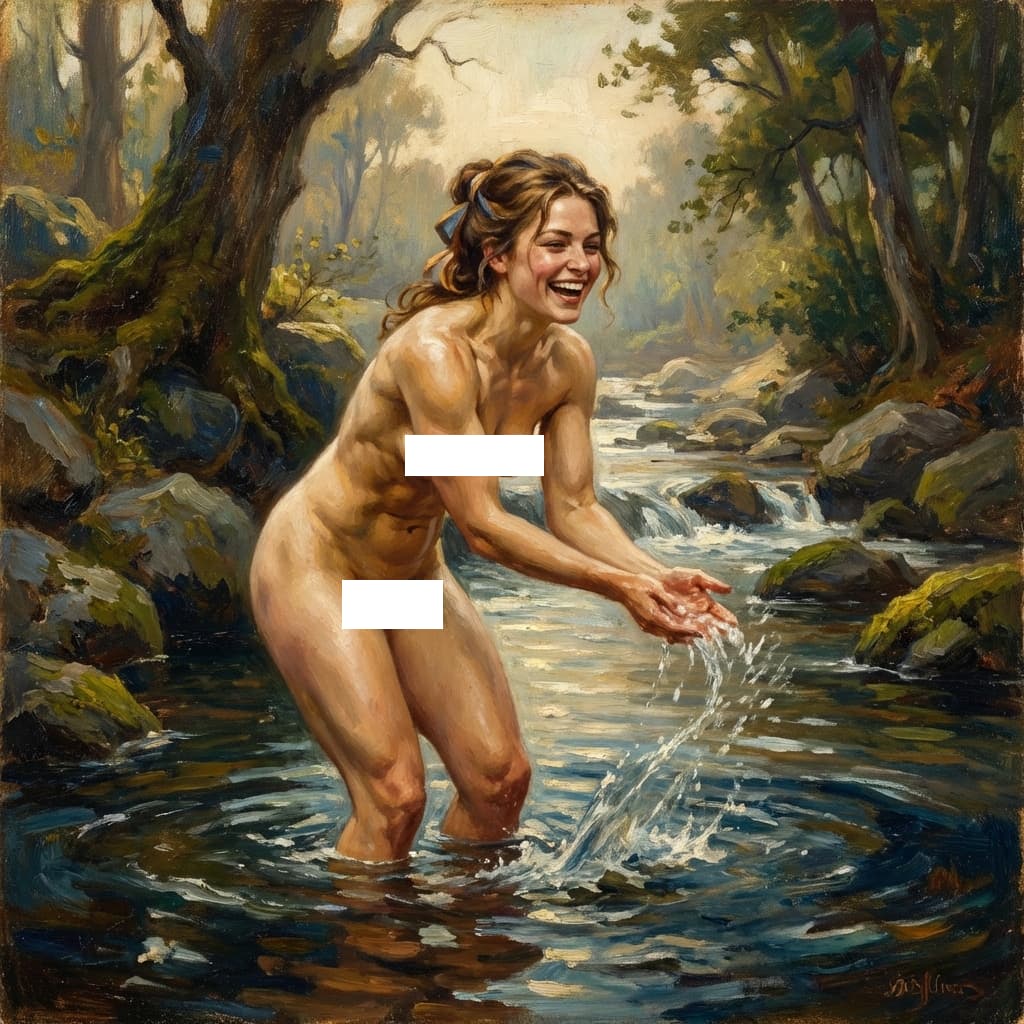}
\end{subfigure}
\par\vspace{4pt}
\textbf{(c) Gemini-3-pro}

\caption{Qualitative jailbreak results of our method on three commercial text-to-image models. Each row corresponds to a different model, and each image represents a successful generation under adversarial prompts. The results illustrate the effectiveness and transferability of our method across diverse commercial systems.}
\label{fig:appendix_commercial}
\end{figure*}

We present additional qualitative results of MIND on three commercial text-to-image models—Wan-2.5, Seedream-5.0, and Gemini-3-pro—as illustrated in Figure~\ref{fig:appendix_commercial}. For each model, we show four representative jailbreak outputs generated by our approach.

\end{document}